\pdfoutput=1

\documentclass[11pt]{article}

\usepackage{ACL2023}

\usepackage{times}
\usepackage{latexsym}
\usepackage[toc,page,header]{appendix}
\usepackage{titletoc}
\usepackage[T1]{fontenc}

\usepackage[utf8]{inputenc}

\usepackage{microtype}

\usepackage{inconsolata}

\usepackage{graphicx}
\usepackage{booktabs}
\usepackage{xltabular}

\usepackage{amsmath,amssymb}
\newcommand{\mycomment}[1]{}
\usepackage{amssymb,amsmath,amsthm}

\newtheorem{theorem}{Theorem}
\newtheorem{lemma}{Lemma}
\usepackage{graphicx}

\usepackage{amsmath,amsfonts,bm}

\def\eqref#1{equation~\ref{#1}}

\def\1{\bm{1}}

\DeclareMathAlphabet{\mathsfit}{\encodingdefault}{\sfdefault}{m}{sl}
\SetMathAlphabet{\mathsfit}{bold}{\encodingdefault}{\sfdefault}{bx}{n}

\DeclareMathOperator*{\argmin}{arg\,min}

\usepackage{subcaption}
\usepackage{xcolor}
\usepackage{enumitem,kantlipsum}

\title{Transductive Learning for Textual Few-Shot Classification \\ in API-based Embedding Models}

\author{Pierre Colombo$^*$$^{2,6}$ \quad Victor Pellegrain$^*$$^{5,6}$, \\ \textbf{Malik Boudiaf}$^{1}$, \quad \textbf{Victor Storchan}$^{3}$, \quad \textbf{Myriam Tami}$^{6}$, \\ \textbf{Ismail Ben Ayed}$^{1}$, \quad  \textbf{Celine Hudelot}$^{6}$, \quad \textbf{Pablo Piantanida}$^{4}$ \\ $^1$\'ETS Montreal, LIVIA, ILLS, Canada \,\ $^2$Equall, Paris, France\,\ \\ $^3$ Mozilla.ai, Paris, France \,\ \quad $^4$ ILLS, MILA, CNRS, CentraleSupélec, Canada \\ \,\ \quad $^5$ IRT SystemX Saclay, France, $^6$ MICS, CentraleSupelec, Universite Paris-Saclay, France\\}

\begin{document}
\maketitle
\def\thefootnote{*}\footnotetext{These authors contributed equally to this work}\def\thefootnote{\arabic{footnote}}

\begin{abstract}

Proprietary and closed APIs are becoming increasingly common to process natural language, and are impacting the practical applications of natural language processing, including few-shot classification. Few-shot classification involves training a model to perform a new classification task with a handful of labeled data. This paper presents three contributions. First, we introduce a scenario where the embedding of a pre-trained model is served through a gated API with compute-cost and data-privacy constraints. Second, we propose a transductive inference, a learning paradigm that has been overlooked by the NLP community. Transductive inference, unlike traditional inductive learning, leverages the statistics of unlabeled data. We also introduce a new parameter-free transductive regularizer based on the Fisher-Rao loss, which can be used on top of the gated API embeddings. This method fully utilizes unlabeled data, does not share any label with the third-party API provider and could serve as a baseline for future research. Third, we propose an improved experimental setting and compile a benchmark of eight datasets involving multiclass classification in four different languages, with up to 151 classes. We evaluate our methods using eight backbone models, along with an episodic evaluation over 1,000 episodes, which demonstrate the superiority of transductive inference over the standard inductive setting.
\end{abstract}

\section{Introduction}

Recent advances in Natural Language Processing (NLP) have been largely driven by the scaling paradigm \citep{kaplan2020scaling,rosenfeld2019constructive}, where larger models with increased parameters have been shown to achieve state-of-the-art results in various NLP tasks \citep{touvron2023llama,radford2019language}. This approach has led to the development of foundation models such as ChatGPT \citep{lehman2023need,kocoń2023chatgpt,brown2020language}, GPT-4 \citep{openai2023gpt4}, GPT-3 \citep{brown2020language}, T5 \citep{raffel2020exploring}, and BERT \citep{devlin2018bert}, which have achieved unprecedented performance in text classification \citep{liu2019roberta}, language modeling, machine translation \citep{fan2021beyond}, and coding tasks \citep{chen2021evaluating}.

\begin{figure*}
    \centering
    \includegraphics[width = 0.8\textwidth]{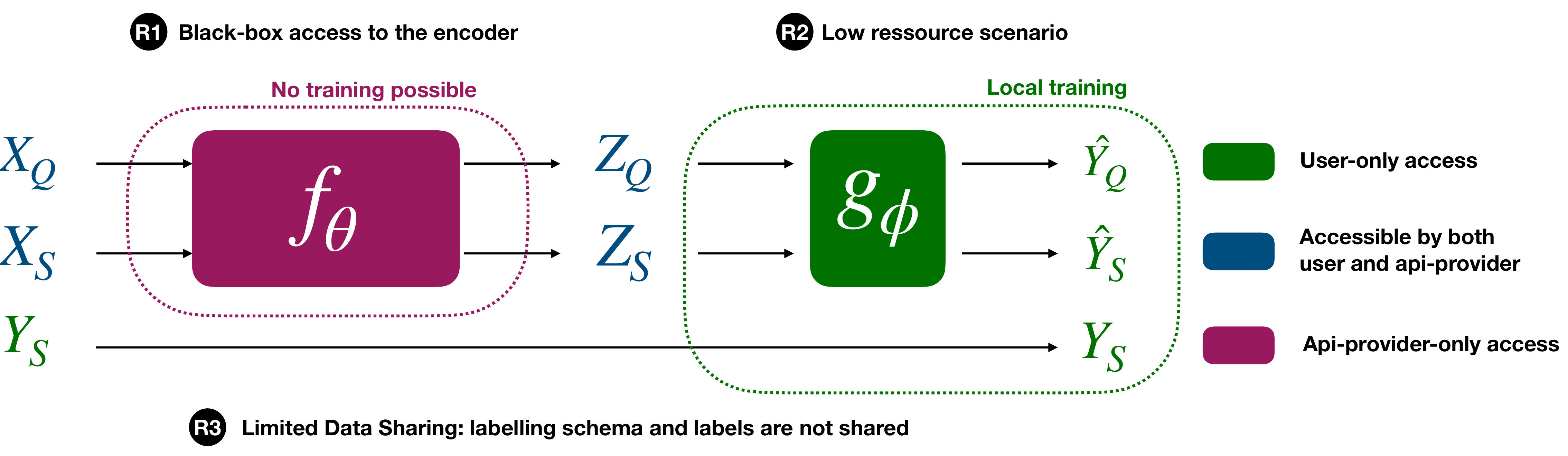}
    \caption{API-based FSL scenario. The black-box API provides embeddings from the pretrained encoder $f_\theta$. The black-box scenario discards existing inductive approaches and in-context learning methods due to the inaccessible of the model's parameters (\textbf{(R1)}) and privacy concerns (\textbf{(R3)}). This scenario, allows tuning a classification head $g_\phi$ (using induction or transduction) at low computational cost \textbf{(R2)} while retaining all support labels locally.}
    \label{fig:setting}
\end{figure*}

Despite the success of the scaling paradigm, significant challenges still exist  especially when the many practical constraints of real-world scenarios have to be met: labeled data can be severely limited (\textit{i.e.,} few-shot scenario \citep{song2022comprehensive,ye2021crossfit}), data privacy is critical for many industries and has become the subject of increasingly many regulatory pieces \citep{eu2020datagovernance,eu2016gdpr}, compute costs need to be optimized \citep{strubell2019energy}. Furthermore, these challenges are made even more complex as stronger foundation models are now available only through APIs (\textit{e.g.}, OpenAI's GPT-3, GPT-4 or ChatGPT, Anthropic's Claude or Google's PaLM \citep{chowdhery2022palm})  which has led to some of their parameters being concealed, presenting new challenges for model adaptation \citep{solaiman2023gradient}. This paper is centered on the fundamental task of few-shot text classification, specifically focusing on cloud-based/API access. Specifically, we formulate three requirements for API-based few-shot learning (FSL) (see  \autoref{fig:setting}):
\\\noindent {\bf (R1)} {\bf Black-box scenario.} We focus on learning from models that are opaquely deployed in production to the end-user, who only has access to the end-point of the encoder, \textit{i.e.}, the resulting text embedding produced by the final layer of the network.
\\\noindent{\bf (R2)} {\bf Low resources / computation time.} AI systems are often required to make rapid predictions at high frequencies in various real-world applications. Therefore, any few-shot classifier used in such scenarios should have a low training and inference time, as well as require minimal computational resources.
\\\noindent {\bf (R3)} {\bf Limited Data Sharing.} When utilizing API models, data sharing becomes a major concern. In the current landscape, providers are increasingly offering less transparent procedures for training their networks. As a result, users prefer sharing as little information as possible, such as labeling schema and annotated data, to safeguard their data privacy. 
\textbf{Shortcomings of Existing Works.} While numerous previous studies have addressed the popular \emph{few-shot} classification setting, to our knowledge no existing line of work adequately satisfies the three API requirements described above. In particular, prompt-based FSL  \citep{schick2020exploiting} and parameter-efficient fine-tuning FSL \citep{houlsby2019parameter} both require access to the model's gradients, while in-context learning scales poorly with the task's size (\textit{e.g} number of shots, number of classes) \citep{chen2021meta,min2021metaicl,min2022rethinking,brown2020language} and requires full data sharing. Instead, we focus on methods that can operate within API-based constraints. 

Under \textbf{R1}, \textbf{R2}, and \textbf{R3} requirements, the standard inductive learning \citep{liu2022few} may be quite limiting. To mitigate the labeled data scarcity while retaining API compliance, we revisit transduction \citep{vapnik1999overview} in the context of textual few-shot classification. Specifically, in the context of FSL, transductive FSL \citep{liu2018learning} advocates leveraging unlabeled test samples of a task as an additional source of information on the underlying task's data distribution in order to better define decision boundaries. Such additional source essentially comes for free in many \textit{offline} applications, including sentiment analysis for customer feedback, legal document classification, or text-based medical diagnosis.

Our findings corroborate recent findings in computer vision \citep{liu2018learning, ziko2020laplacian, lichtenstein2020tafssl, boudiaf2020information, hu2021leveraging}, that substantial gains can be obtained from using transduction over induction, opening new avenue of research for the NLP community. However, the transductive gain comes at the cost of introducing additional hyperparameters, and carefully tuning them. Motivated by Occam's razor principle, we propose a novel hyperparameter-free transductive regularizer based on Fisher-Rao distances and demonstrate the strongest predictive performances across various benchmarks and models while keeping hyperparameter tuning minimal. We believe that this parameter-free transductive regularizer can serve as a baseline for future research.

\subsection*{Contributions}
In this paper, we make several contributions to the field of textual FSL. Precisely, our contributions are threefold:
\\\noindent\textbf{A new textual few-shot scenario:} We present a new scenario for FSL using textual API-based models that accurately capture real-world constraints. Our new scenario opens up new research avenues and opportunities to address the challenges associated with FSL using API-based models, paving the way for improved performance in practical applications.
\\\noindent\textbf{A novel transductive baseline.} Our paper proposes a transductive FSL algorithm that utilizes a novel parameter-free Fisher-Rao-based loss. By leveraging only the network's embedding {\bf (R1)}, our approach enables fast and efficient predictions {\bf (R2)} without the need to share the labeling schema or the labels of few-shot examples making it compliant with {\bf (R3)}. This innovative method marks a significant step forward in the field of FSL.
\\\noindent\textbf{A truly improved experimental setting.} Previous studies on textual few-shot classification \citep{schick2022true,schick2020s,mahabadi2022perfect,tam2021improving,gao2020making} have predominantly assessed their algorithms on classification tasks with a restricted number of labels (typically less than five). We take a step forward and create a benchmark that is more representative of real-world scenarios. Our benchmark relies on a total of eight datasets, covering multiclass classification tasks with up to 151 classes, across four different languages.  Moreover, we further enhanced the evaluation process by not only considering 10 classifiers trained with 10 different seeds \citep{logan2021cutting,mahabadi2022perfect}, but also by relying on episodic evaluation on 1,000 episodes \citep{hospedales2021meta}. Our results clearly demonstrate the superiority of transductive methods.
\section{Related Work}
\subsection{Few-shot learning in NLP}
Numerous studies have tackled the task of FSL in Natural Language Processing (NLP) by utilizing pre-trained language models \citep{devlin2018bert,liu2019roberta,radford2019language,yang2019xlnet}. These methods can be classified into three major categories: prompt-based, parameter-efficient tuning, and in-context learning.
\\\noindent \textbf{Prompt-based FSL}:  Prompt-based FSL involves the use of natural language prompts or templates to guide the model to perform a specific task \citep{ding2021openprompt,liu2023pre}. For example, the seminal work \citep{schick2020exploiting} proposed a model called PET, which uses a pre-defined set of prompts to perform various NLP tasks as text classification. They also impose a choice of a verbalizer which highly impacts the classification performances \citep{cui2022prototypical,hu2021knowledgeable}. However, recent studies have questioned the benefits of prompt-based learning due to the high variability in performance caused by the choice of prompt \citep{liu2022few}. To address this issue, researchers have proposed prompt tuning which involves a few learnable parameters in addition to the prompt \citep{lester2021power}. Nevertheless, these approaches face limitations when learning from API: (i) encoder access for gradient computation is infeasible (as in \textbf{R1}), (ii) prompting requires to send data and label which raises privacy concerns (as in \textbf{R3}), and (iii) labeling new points is time-consuming (see in \textbf{R3}) and expensive due to the need to send all shots for each input token\footnote{The cost of API queries is determined by the number of input tokens that are transmitted.}.
\\\noindent\textbf{Parameter-efficient fine-tuning.} These methods, such as adapters \citep{houlsby2019parameter,pfeiffer2020adapterhub}, keep most of the model's parameters fixed during training and only update small feed-forward networks that are inserted within the larger model architecture. A recent example is T-FEW \citep{liu2022few}, which adds learned vectors that rescale the network's internal activations. Additionally, it requires a set of manually created prompts for each dataset making it hard to use in practice. Relying on parameter-efficient fine-tuning methods with an API is not possible due to the need to compute gradients of the encoder (as per \textbf{R1}) and the requirement to send both the labeling schema and the labels, which violates \textbf{R3}.
\\\noindent\textbf{In Context Learning (ICL).} In-context learning models are models that utilize input-to-output training examples as prompts to make predictions, without any parameter updates \cite{wei2022chain}. These models, such as \texttt{text-davinci}, rely solely on the provided examples to generate predictions, without any additional training. However, a significant drawback of this approach is that the user must supply the input, label examples, and task description, which becomes prohibitively expensive when the number of classes or shots increases, is slow \citep{liu2022few} (\textbf{R2}) and raises data privacy concerns (as highlighted in \textbf{R3}). Additionally, the inability to reuse text embeddings for new tasks or with new labels without querying the model's API limits practicality and scalability, making reusable encoding unfeasible for in-context learning models\footnote{Furthermore, as the number of considered classes increases, the fixed size of the transformer limits the number of possible shots that can be fed to the model. Previous studies have often neglected this limitation by focusing on a few numbers of labels.}.
\\\noindent \textbf{Meta-learning.} Meta-learning approaches have for quite long stood as the \textit{de-facto} paradigm for FSL (\cite{snell2017prototypical, Rusu-2018-embedding-optimization, sung2018learning, lee2019meta, raghu2019rapid,sun2019meta}). In meta-learning, the objective is to provide the model with the intrinsic ability to learn in a data-efficient manner. For instance, MAML (\cite{finn2017model, antoniou2018train}), arguably the most popular meta-learning method, tries to train a model such that it can be fine-tuned end-to-end using only a few supervised samples while retaining high generalization ability. Unlike the three previous lines of work, meta-learning methods operate by modifying the pre-training procedure and therefore assume access to both the training data and the model, which wholly breaks both \textbf{R1} and \textbf{R3}.

\subsection{Inductive vs transductive learning}

Learning an inductive classifier on embeddings generated by an API-based model, as proposed by \citep{snell2017prototypical}, is a common baseline for performing FSL. This approach is prevalent in NLP, where a parametric model is trained on data to infer general rules that are applied to label new, unseen data (known as inductive learning \citep{vapnik1999overview}). However, in FSL scenarios with limited labeled data, this approach can be highly ambiguous and lead to poor generalization.

Transduction offers an attractive alternative to inductive learning \citep{sain1996nature}. Unlike inductive learning, which infers general rules from training data, transduction involves finding rules that work specifically for the unlabeled test data. By utilizing more data, such as unlabeled test instances, and aiming for a more localized rule rather than a general one, transductive learning has shown promise and practical benefits in computer vision \citep{boudiaf2020information,boudiaf2021few,ziko2020laplacian}. Transductive methods yield substantially better performance than their inductive counterparts by leveraging the statistics of the query set \citep{dhillon2019baseline}. However, this approach has not yet been explored in the context of textual data.

\section{API-based Few-shot Learning}

\subsection{Problem Statement}\label{ssec:pb_str}

Let $\Omega$ be the considered vocabulary, we denote $\Omega^*$ its Kleene closure. The Kleene closure corresponds to sequences of arbitrary size written with tokens in $\Omega$, \textit{i.e.}, $\Omega^* = \overset{\infty}{\underset{i = 0}{\bigcup}} \Omega^i$. Given an input space $\mathcal{X}$ with $\mathcal{X} \subseteq \Omega^*$ and a latent space $\mathcal{Z}$, we consider a pre-trained backbone model $f_\theta : \mathcal{X} \rightarrow \mathcal{Z} = \mathcal{R}^d$, where $\theta \in \Theta$ represents the parameters of the encoder and $d$ is the embedding dimension size. In the API-based setting, we assume that we are unable to access the exact structure of $f_\theta$ as mentioned in \textbf{R1}. However, we do have access to the last encoder embedding which is available for our use (see \textbf{R1}).

The objective of few-shot classification is to learn a classifier from limited labeled data and generalize it to new, unseen tasks or classes. To accomplish this, randomly
sampled few-shot tasks are created from a test dataset $\mathcal{D}_{test}:=\{(x_i, y_i)\}_{i=1}^{N_{test}}$ that has a set of unseen classes $\mathcal{Y}_{test}$. Each task involves a few labeled examples from $K$ different classes chosen at random among $\mathcal{Y}_{test}$. These labeled examples constitute the support set $S = \left\{x_i,y_i\right\}_{i \in \mathcal{I}_S}$, with a size of $|S|=N_S\times K$. Additionally, each task has an unlabeled query set $Q= \left\{x_i\right\}_{i \in \mathcal{I}_Q}$ composed of $|Q|=N_Q\times K$ unseen examples from each of the $K$ classes. $\mathcal{I}_S$ and $\mathcal{I}_Q$ represent the drawn indices during the sampling process for support set and query set, respectively. Pre-trained models use few-shot techniques and the labeled support sets to adapt to the tasks at hand and are evaluated based on their performances on the unlabeled query sets.

\textit{Remark} Setting the values of $N$ and $K$ in textual FSL is not standardized, as discussed in \autoref{ssec:pb_str}. Therefore, in all of our experiments, we have relied on setting $(N,K) \in \{5,10\}^2$.

\subsection{Proposed Transductive Method}\label{ssec:model_trans}
NLP few-shot classifiers rely only on inductive inference, while computer vision has shown significant performance improvements using transductive inference for FSL. Transductive inference succeeds in FSL because it jointly classifies all unlabeled query samples of a single task, leading to more efficient and accurate classification compared to inductive methods that classify one sample at a time. Let us begin by introducing some basic notation and definitions before introducing our new  transductive loss based on the Fisher-Rao distance.

In the API-based few-shot classification setting, our goal is to train a classification head $g_{\phi}: \mathcal{Z} \to \mathbb{R}^K$ that maps the feature representations to the posterior distribution space for making predictions. To simplify the equations for the rest of the paper, we use the following notations for the posterior predictions of each $i \in \mathcal{I}_S\cup \mathcal{I}_Q$ and for the class marginals within $Q$: $p_{ik} = g_{\phi}(f_{\theta}(x_i))_k = \mathbb{P}(Y=k|X=x_i; \theta, \phi)$ and $ \widehat{p}_k = \frac{1}{|Q|}\sum_{i \in \mathcal{I}_Q} p_{ik} =\mathbb{P}(Y_Q=k; \theta, \phi)$
 where $X$ and $Y$ are the r.v.s associated with the raw features and labels, respectively, and where $Y_Q$ means restriction of the r.v. $Y$ to set $Q$. 

For training the classification head in the transductive setting, prior research aims at finding $\phi$ such that $\phi = \argmin \text{CE} - \lambda \times R_Q$\footnote{$\lambda$ is set to 1 in all the experiments.}, with CE$:= - \frac{1}{|S|}\sum_{i\in \mathcal{I}_S}\sum_{k=1}^K y_{ik}\log(p_{ik})$ being the cross-entropy supervision on the support set (in which $y_{ik}$ is the $k^{\text{th}}$ coordinate of the one-hot encoded label vector associated to sample $i$) and ${R}_Q$ being a transductive loss on the query set $Q$. 

Note that this transductive regularization has been proposed in the literature based on the InfoMax principle \citep{cardoso1997infomax,linsker1988self}, and the inductive loss can be found by setting $\lambda=0$. In what follows, we review the regularizers introduced in previous work.


\textbf{Entropic Minimization (H)} An effective regularizer for transductive FSL can be derived from the field of semi-supervised learning, drawing inspiration from the approach introduced in \citep{grandvalet2004semi}. This regularizer, proposed in \citep{dhillon2019baseline}, utilizes the conditional Shannon Entropy \citep{cover1999elements} of forecast results from query samples during testing to enhance model generalization. Formally:
\begin{equation}\label{eq:entropy}
    R_Q^H =  \frac{1}{|Q|}\sum_{i \in \mathcal{I}_Q}\sum_{k=1}^K p_{ik}\log(p_{ik}).
\end{equation}
\textbf{Mutual Information Maximization (I)} A promising alternative to entropic minimization for addressing the challenges of transductive FSL is to adopt the Info-max principle. \citep{boudiaf2020information} extended this idea, introduced in  \citep{Hu2017-imsat}, and propose as regularizer a surrogate of the mutual-information $R_Q^{I}(\alpha)=$: 
\begin{equation}
\label{eq:TIM}
\hspace{-.1cm}-\sum_{k=1}^K \hat{p}_k \log \hat{p}_k + \alpha \frac{1}{|Q|}\sum_{i \in \mathcal{I}_Q}\sum_{k=1}^K p_{ik}\log(p_{ik}).
\end{equation} 


\textbf{Limitation of existing strategies}: Despite its effectiveness, the previous method has a few limitations that should be taken into account. One of these limitations is the need to fine-tune the weight of different entropies using the hyperparameter $\alpha$. This parameter-tuning process can be time-consuming and may require extensive experimentation to achieve optimal results. Additionally, recent studies have shown that relying solely on the first Entropic term, which corresponds to the Entropic minimization scenario in Equation \ref{eq:entropy}, can lead to suboptimal performance in FSL.

\subsection{A Fisher-Rao Based Regularizer}

In the FSL scenario, minimizing parameter tuning is crucial. Motivated by this, in this section, we introduce a new parameter-free transductive regularizer that fits into the InfoMax framework. Additionally, our loss inherits the attractive properties of the Fisher-Rao distance between soft-predictions $\mathbf{q}:=(q_1,\dots,q_K)$ and $\mathbf{p}:=(p_1,\dots,p_K)$, which is given by~\citep{9773978}:
\begin{equation}  
d_{\textrm{FR}}(\mathbf{q},\mathbf{p}) :=
2 \arccos \left( \sum_{k=1}^K \sqrt{q_k \times p_k} \right). 
\label{eq:fisher_rao_multiclass}
\end{equation}
The proposed transductive regularizer denoted by $R^{\textrm{FR}}_Q$, for each single few-shot task, can be described as measuring the Fisher-Rao distance between pairs of query samples: 
\begin{align}
\label{eq:MI-FR}
&R^{\textrm{FR}}_Q:= \frac{1}{|Q|}\sum_{i\in \mathcal{I}_Q} -\log \sum_{j\in \mathcal{I}_Q} \sum_{k=1}^K  \sqrt{\displaystyle p_{ik} \times p_{jk}} \\ &= \frac{1}{|Q|}\sum_{i\in \mathcal{I}_Q} -\log \sum_{j\in \mathcal{I}_Q} \cos\left(\frac{d_{\textrm{FR}}(\mathbf{p}_i,\mathbf{p}_j)}{2}\right), 
\end{align} 
where $d_{\textrm{FR}}(\mathbf{p}_i,\mathbf{p}_j)$ is the Fisher-Rao distance between pairs of soft-predictions $(\mathbf{p}_i,\mathbf{p}_j)$. Furthermore, it is shown that expression (\ref{eq:MI-FR}) yields a surrogate of the Mutual Information as shown by the following proposition. This result to the best of our knowledge is new, as far as we can tell. 

\begin{theorem}\label{th:one}(Fisher-Rao as a surrogate to maximize Mutual Information) 
Let $(\mathbf{q}_i)_{i\in \mathcal{I}_Q}$ be a collection of soft predictions corresponding to the query samples. Then, it holds that $\forall$ $0 \leq  \alpha\leq 1$: 
\begin{align}\label{eq:kl_bound1}
    \begin{split}
     R^{\textrm{FR}}_Q +\log |Q| \leq R_Q^{I}( 1 )\leq R_Q^{I}( \alpha ),
        \end{split}
\end{align}
\end{theorem}

\textit{Proof:} 
Further details are relegated to \autoref{sec:proofs}.

\textit{Advantage of $R^{\textrm{FR}}_Q$ over $R^{I}_Q(\alpha)$:} Similarly to $R^{I}_Q(\alpha)$, $R^{\textrm{FR}}_Q$ can be exploited to maximize the Mutual Information. However, $R^{\textrm{FR}}_Q$ is parameter-free and thus, it does not require tuning $\alpha$.

\subsection{Additional Few-shot Inductive Baseline}
In addition to the transductive methods of \autoref{ssec:model_trans}, we will explore three additional inductive methods for few-shot classification: prototypical networks, linear probing, and a semi-supervised classifier. 

\textbf{Prototypical Networks (PT)} PT learn a metric space where the distance between two points corresponds to their degree of similarity. During inference, the distance between the query example and each class prototype is computed, and the predicted label is the class with the closest prototype. PT has been widely used in NLP and is considered as a strong baseline \citep{snell2017prototypical,sun2019hierarchical,gao2019hybrid}.

\textbf{Linear Probing (CE)} Fine-tuning a linear head on top of a pretrained model is a popular approach to learn a classifier for classification tasks and was originally proposed in \citep{devlin2018bert}. 

\textbf{Semi-supervised Baselines (SSL)}. We additionally propose two semi-supervised baselines following two steps. In the first step, a classifier is trained using the support set $\mathcal{S}$ and used to label $\mathcal{Q}$. In the second step, the final classifier is trained on both  $\mathcal{S}$  and  $\mathcal{Q}$ with the pseudo label obtained from the first step.


\begin{table}
\caption{Datasets Statistics.}\label{tab:dataset}\vspace{-.2cm}
\centering
\begin{tabular}{cc}\hline
  Dataset   & Classes (K) \\\hline
   Tweet  & 20 \\
   Emotion   & 25 \\
   Amazon  & 30 \\
   B77& 77 \\
      Clinc & 151 \\\hline
\end{tabular}\vspace{-.5cm}
\end{table}

\section{An Enhanced Experimental Setting}
\subsection{Datasets}
 Benchmarking the performance of FSL methods on diverse sets of datasets is critical to evaluate their generalization capabilities in a robust manner as well as their potential for real-world applications. Previous work on FSL \citep{karimi2022perfect,perez2021true} mainly focuses on datasets with a reduced number of classes (\textit{i.e.}, $K < 5$). Motivated by practical considerations we choose to build a new benchmark composed of datasets with a larger number of classes. Specifically, we choose Go Emotion \citep{demszky2020goemotions}, Tweet Eval \citep{barbieri2020tweeteval}, Clinc \citep{larson-etal-2019-evaluation}, Banking \citep{casanueva-etal-2020-efficient} and the Multilingual Amazon Reviews Corpus \citep{keung-etal-2020-multilingual}. These datasets cover a wide range of text classification scenarios and are of various difficulty\footnote{Datasets are available in Dataset \citep{lhoest2021datasets}}. A summary of the datasets used can be found in \autoref{tab:dataset}.

\subsection{Model Choice}
The selection of an appropriate backbone model is a critical factor in achieving high performance in few-shot NLP tasks. To ensure the validity and robustness of our findings, we have included a diverse range of transformer-based backbone models in our study, including
\begin{enumerate}[wide, labelwidth=!, labelindent=0pt]
\item \textit{Three different sizes of RoBERTa based models} \citep{liu2019roberta}. Similar to BERT, RoBERTa is pretrained using the closed task \citep{taylor1953cloze}. We consider two different sizes of the RoBERTa model, namely RoBERTa (B) with 124M parameters and RoBERTa (L) with 355M parameters and DistilRoBERTa, a lighter version of RoBERTa trained through a distillation process \citep{Hinton2015DistillingTK}, for a total of 82M parameters. 
\item \textit{Three sentence-transformers encoder} \citep{reimers-gurevych-2019-sentence}. Following \citep{muennighoff2022mteb}, we consider MPNET-base \citep{song2020mpnet}, MiniLM \citep{wang2020minilm}, and Albert Small V2 \citep{lan2019albert}.
\item \textit{Multilingual models.} To address realistic multilingual scenarios, we rely on three sizes of XLM-RoBERTa \citep{conneau-etal-2020-unsupervised,conneau2019unsupervised}: base (B), large (L) and XL (XL).
\item \textit{\texttt{text-davinci} model}: to mimic the typical setting of API-based models, we also conduct experiments on \texttt{text-davinci}, only accessible through OpenAI's API. 
\end{enumerate}

\subsection{Evaluation Framework}
Prior research in textual FSL typically involves sampling a low number of tasks, typically less than 10, of each dataset. In contrast, we utilize an episodic learning framework that generates a large number of N-shots K-ways tasks. This framework has gained popularity through inductive meta-learning approaches, such as those proposed by \citep{finn-17-maml, snell2017prototypical,vinyals-2016-matching,sung-2018-compare,Mishra2017ASN,Rusu-2018-embedding-optimization,oreshkin-2018-tadam}, as it mimics the few-shot environment during evaluation and improves model robustness and generalization. In this context, episodic training implies that a different model is initialized for each generated few-shot task, and all tasks are compiled independently in parallel. This approach allows to the computation of more reliable performance statistics by evaluating the generalization capabilities of each method on a more diverse set of tasks. 
To account for the model's generalization ability, we average the results for each dataset over 1000 episodes, with the N considered classes varying in every episode. For each experiment, we consider the F1-Score.


\vspace{-0.2cm}
\section{Experiments}\label{sec:experiment}
\subsection{Case Study of \texttt{text-davinci}}
\begin{table}
\caption{Aggregated performance over K, N, the different datasets for \texttt{text-davinci}.  $|x|$  stands for the averaged input length.}\label{tab:davinci}\vspace{-.2cm}
\resizebox{0.4\textwidth}{!}{
\begin{tabular}{lrrrrc}\hline
N-shots &     \multicolumn{2}{c}{10} &     \multicolumn{2}{c}{5} &  $|x|$ \\
K-ways &     10 &      5 &     10 &      5 & \\\hline
FR     &  \textbf{69.83} &  \textbf{77.46} &  \textbf{66.70} &  \textbf{75.03} & 14.2\\
H      &  10.00 &  20.00 &  10.01 &  20.04 &14.2\\
I      &  \textbf{68.38} &  \textbf{75.82} &  \textbf{65.15} &  \textbf{73.06} &14.2\\\hline
CE     &  68.21 &  75.47 &  64.92 &  72.70 &14.2\\
PT     &  67.95 &  75.41 &  64.60 &  72.50 & 14.2\\
SSL    &  68.27 &  75.55 &  64.99 &  72.75 & 14.2 \\\hline
ICL    &  68.9 &  76.24 &  65.2 &  74.3 & 900 \\\hline
\end{tabular}
}\vspace{-.5cm}
\end{table}
In this experiment, we investigate the performance of \texttt{text-davinci} in both its language model and embedding-based model forms. We assess its classification capabilities using the aforementioned baseline and explore the language model's performance when applied in an in-context learning (ICL) setup with prompting.

\textbf{Takeaways}. From \autoref{tab:davinci}, we observe that SSL performs comparably to CE, which is simpler to use and will be considered as the baseline in the next part of our study. Although ICL slightly outperforms CE, its implementation comes at a significant cost. In ICL, each class requires N shots, forcing the user to send a long input query with additional instructions. This query length becomes prohibitive as the number of classes increases, and on average, it is 58 times longer than using the embedding base API in our benchmark. The lengthy input and ICL approach make it time-consuming for generation (violating \textbf{R1}), require the user to provide labels (violating \textbf{R2}), and prevent the reuse of embeddings for future use (\textit{e.g.}, retrieval, clustering). Additionally, ICL is 60 times more expensive than CE. Thus, we will discard ICL for the subsequent part of this study.


\subsection{Overall Results}
\label{subsec:global}
\textbf{Global results:}
To evaluate the effectiveness of various few-shot methods, we conducted a comprehensive analysis of their classification performance across all datasets, all backbones, and all considered N-shots/K-ways scenarios. Results are reported in \autoref{tab:global}. \emph{An interesting observation is that transductive approaches I and FR outperform their inductive counterparts (CE and PT).} Notably, we found that vanilla entropy minimization, which solely relies on H, consistently underperforms in all considered scenarios. Our analysis revealed that FR surpasses traditional fine-tuning based on cross-entropy by a margin of 3.7\%.

\begin{table}
\caption{Aggregated performance over K,N, the different datasets and considered backbone.}\label{tab:global}
\centering
\resizebox{0.4\textwidth}{!}{\begin{tabular}{lrrrr} \hline
N-shots &   \multicolumn{2}{c}{10}  &     \multicolumn{2}{c}{5}  \\
 K-ways &     10 &      5 &     10 &      5 \\ \hline
  FR     &  \textbf{52.09} &  \textbf{61.99} &  \textbf{48.71} &  \textbf{56.55} \\ 
   I      &  \underline{50.07} &  \underline{59.17} &  \underline{46.42} &  \underline{55.74} \\ 
    H      &  15.07 &  27.39 &  15.33 &  25.84 \\\hline
 CE     &  48.31 &  56.87 &  45.27 &  53.94 \\
  SSL     &  50.39&	58.78	&47.33	&55.85 \\
 PT     &  47.29 &  56.05 &  44.32 &  53.20 \\\hline
 \end{tabular}}
\end{table}

\begin{figure}
\subfloat{{\includegraphics[width=7.2cm]{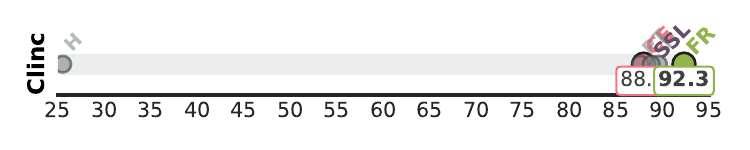} }} \\ \subfloat{{\includegraphics[width=7.2cm]{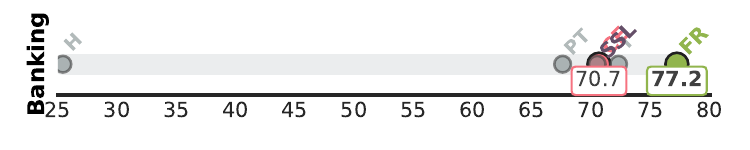} }} \\ \subfloat{{\includegraphics[width=7.2cm]{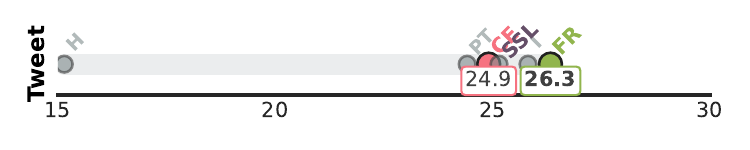} }} \\ \subfloat{{\includegraphics[width=7.2cm]{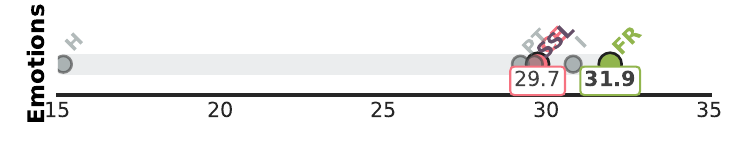} }}\caption{Performance on the monolingual datasets.}%
    \label{fig:monolingual}\vspace{-.5cm}
\end{figure}

\textbf{Mono-lingual experiment}: In order to thoroughly analyze the performance of each method, we conducted a per-dataset study, beginning with a focus on the mono-lingual datasets. \autoref{fig:monolingual} reveals that the global trends observed in \autoref{tab:global} remain consistent across datasets of varying difficulty levels. 
Notably, we observed consistent improvements achieved by transductive regularizers (such as I or FR) over CE. However, the relative improvement is highly dependent on the specific dataset being evaluated. Specifically, FR achieves +6.5\% F1-score on Banking, but only a shy +1.5\% on Tweet. A strong baseline generally suggests highly discriminative features for the task, and therefore a strong upside in leveraging additional unlabeled features, and vice versa. Therefore, we hypothesize that the potential gains to be obtained through transduction correlate with the baseline's performance.\footnote{Additional multilingual results (\textit{i.e.}, on es, de, fr) can be found on \autoref{ssec:additionnal_multilingua}. They exhibit the same behavior.}

\begin{figure*}[ht]
    \centering
    \subfloat{{\includegraphics[width=0.5\textwidth]{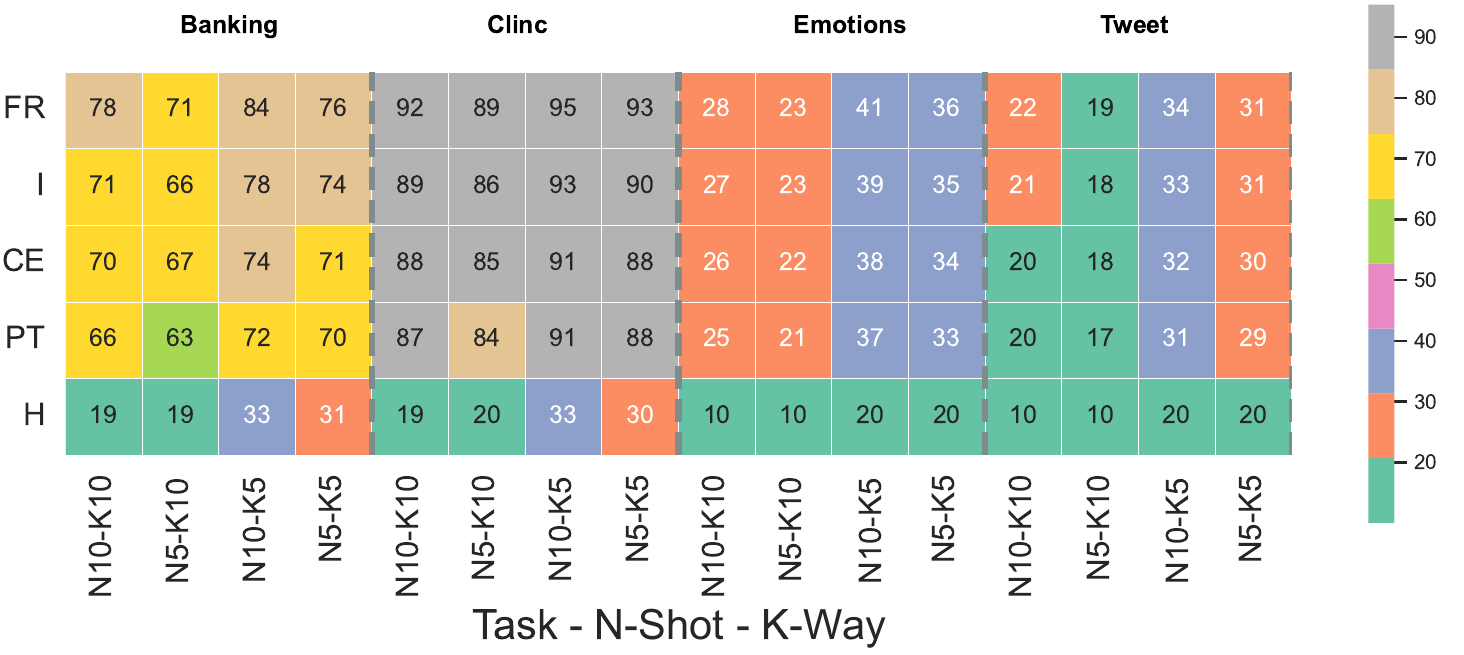}} }%
    \subfloat{{\includegraphics[width=0.5\textwidth]{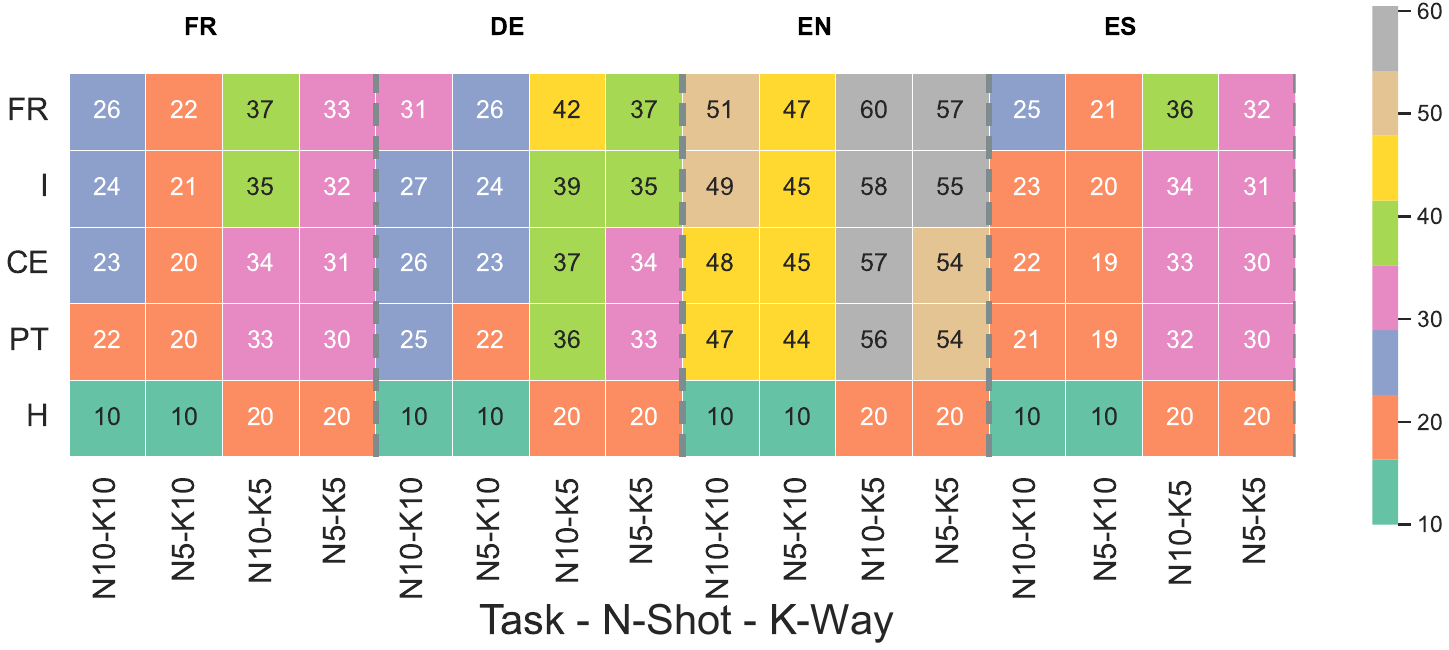}} } 
    \caption{The effect of ways and shots on test performance on monolingual (left) and multilingual (right) datasets.}
    \label{fig:n_way_k_shot}\vspace{-.2cm}
\end{figure*}

\subsection{Study Under Different Data-Regime}

In this experiment, we investigated the performance of different loss functions under varying conditions of 'ways' and 'shots'. As shown in \autoref{fig:n_way_k_shot}, we observed that increasing the number of classes ('ways') led to a decrease in F1 while increasing the number of examples per class ('shots') led to an improvement in F1. This can be explained by the fact that having more data enables the classifier to better discern the characteristics of each class. 

Interestingly, the relationship between the number of shots and classification F1  may not be the same for all classes or all loss functions. \autoref{fig:n_way_k_shot} shows that different loss functions (e.g. FR on banking) benefited greatly from adding a few shots, while others did not show as much improvement. However, this variability is dependent on the specific dataset and language being used, as different classes may have different levels of complexity and variability, and some may be inherently easier or harder to classify than others.

\subsection{Ablation Study On Backbones}

\begin{figure}
    \subfloat{{\includegraphics[width=0.22\textwidth]{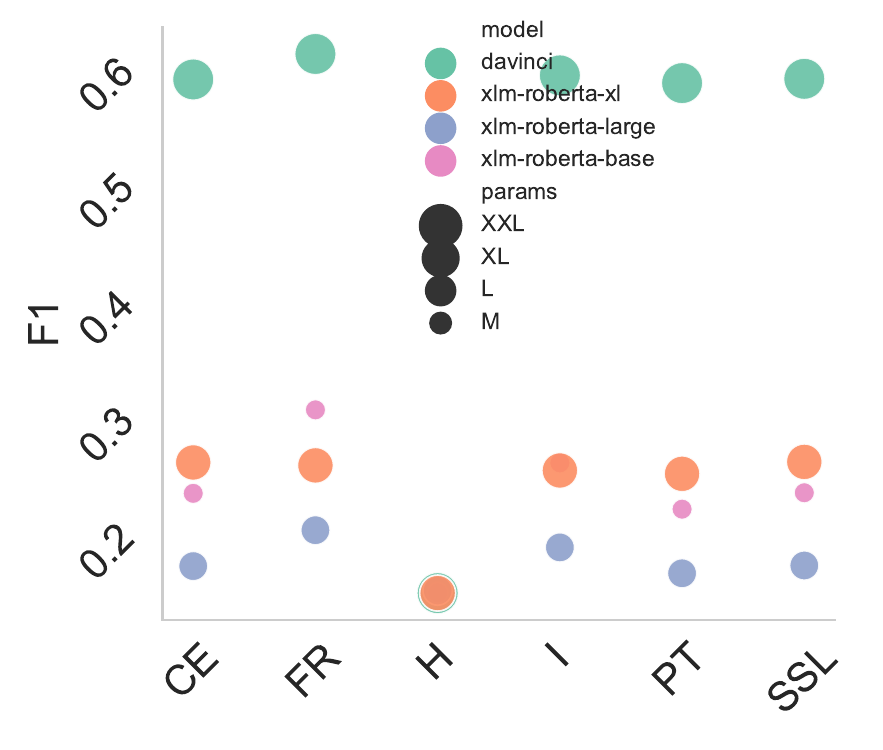}} }%
    \subfloat{{\includegraphics[width=0.22\textwidth]{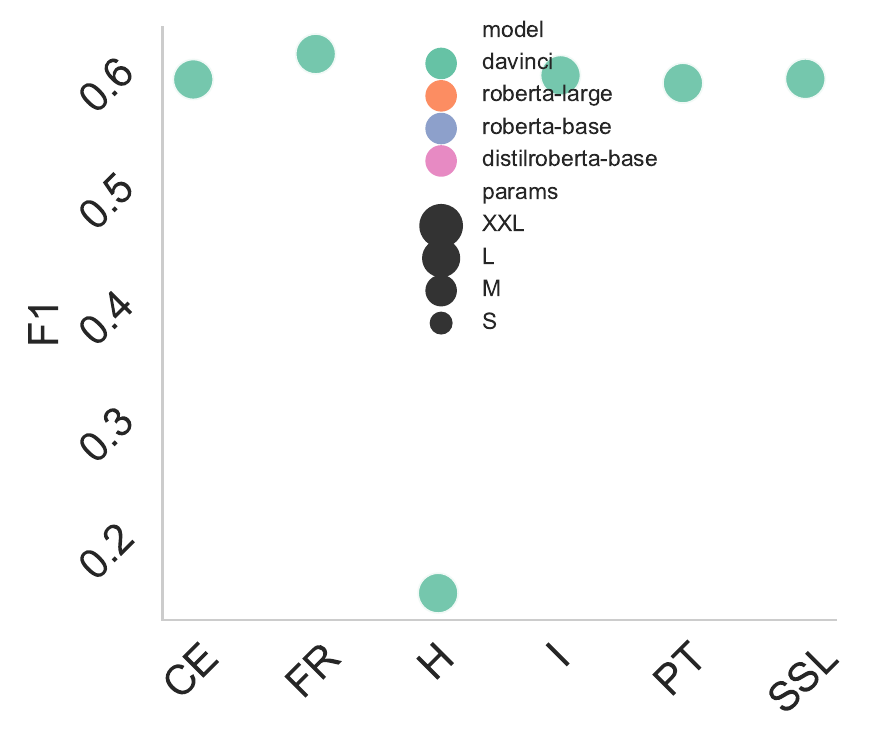}} } 
      \caption{Impact of model size.}
    \label{fig:scaling_laws}\vspace{-.5cm}
\end{figure}

In this experiment, we examined how different loss functions perform when increasing the number of parameters in various models. The results, presented in \autoref{fig:scaling_laws}, show the average performance across the experiments and are organized by the loss function. We observed an \emph{inverse scaling law} for both the RoBERTa and XLM-RoBERTa family of models, where increasing the number of parameters led to a decrease in performance for the losses tested. However, within the same family, we observe that the superiority of FR remains consistent.  An interesting finding from \autoref{fig:scaling_laws} is that the transductive regularization technique using FR outperforms other methods on \texttt{text-davinci}. This highlights the effectiveness of FR in improving the performance of the model and suggests that transductive regularization may be a promising approach for optimizing language models.

\subsection{Practical Considerations}

In this experiment, we adopt a practical standpoint and aim to evaluate the effectiveness of an API model, specifically \texttt{text-davinci}. In \autoref{tab:speed}, we report the training speed of one episode on a MAC with CPU. Overall, we observed that the transductive loss is slower as it necessitates the computation of the loss on the query set, whereas PT is faster as it does not involve any optimization. Furthermore, we note that FR is comparable in speed to I. To provide a better understanding of these results, we can compare our method with existing approaches (in the light of \textbf{R2}). For instance, PET \citep{schick2020exploiting} entails a training time of 20 minutes on A100, while ADAPET \citep{tam2021improving} necessitates 10 minutes on the same hardware.



\begin{table}[]
\centering
    \caption{Training time for 1 episode on a M1-CPU.}\label{tab:speed}\vspace{-.2cm}
    \centering
    \begin{tabular}{cc}\hline
        Loss  & CPU Time \\\hline
       CE  & 0.45s \\
        FR &  0.83s\\
        H &  0.75s\\
        I &  0.83s\\
        PT &  0.01s\\
        SSL &  0.80s\\\hline
    \end{tabular}\vspace{-.5cm}
\end{table}



\section{Conclusions}
This paper presents a novel FSL framework that utilizes API models while meeting critical constraints of real-world applications (i.e., \textbf{R1}, \textbf{R2}, \textbf{R3}). This approach is particularly appealing as it shifts the computational requirements (\textbf{R2}), eliminating the need for heavy computations for the user and reducing the cost of embedding. 
To provide a better understanding, embedding over 400k sequences cost as low as 7 dollars. In this scenario, our research highlights the potential of transductive losses, which have previously been disregarded by the NLP community. A candidate loss is the Fisher-Rao distance which is parameter-free and could serve as a simple baseline in the future.

\newpage
\section{Limitations} 
We are optimistic that our research will have a positive impact on society. Nonetheless, it is essential to acknowledge the limitations of API-based few-shot classification models despite their promising results in various tasks. Firstly, the performance of the introduced methods is heavily dependent on the quality of available API models. If the API models do not provide sufficient information or lack diversity, the introduced methods may struggle to accurately classify input texts. Secondly, the black-box nature of the backbone limits the interpretability of API-based few-shot classification methods, which may hinder their adoption. Ultimately, the aim of this work is to establish a baseline for future research on transductive inference. As a result, not all existing transductive methods are compared in this study.

\section*{Acknowledgements}
This work was performed using HPC resources from GENCI-IDRIS (Grants 2022- AD01101838, 2023-103256 and 2023-101838).

\newpage
\bibliography{tmlr}

\begin{thebibliography}{105}
\expandafter\ifx\csname natexlab\endcsname\relax\def\natexlab#1{#1}\fi

\bibitem[{Antoniou et~al.(2018)Antoniou, Edwards, and
  Storkey}]{antoniou2018train}
Antreas Antoniou, Harrison Edwards, and Amos Storkey. 2018.
\newblock How to train your maml.
\newblock \emph{arXiv preprint arXiv:1810.09502}.

\bibitem[{Barbieri et~al.(2020)Barbieri, Camacho-Collados, Espinosa-Anke, and
  Neves}]{barbieri2020tweeteval}
Francesco Barbieri, Jose Camacho-Collados, Luis Espinosa-Anke, and Leonardo
  Neves. 2020.
\newblock {TweetEval:Unified Benchmark and Comparative Evaluation for Tweet
  Classification}.
\newblock In \emph{Proceedings of Findings of EMNLP}.

\bibitem[{Boudiaf et~al.(2021)Boudiaf, Kervadec, Masud, Piantanida, Ben~Ayed,
  and Dolz}]{boudiaf2021few}
Malik Boudiaf, Hoel Kervadec, Ziko~Imtiaz Masud, Pablo Piantanida, Ismail
  Ben~Ayed, and Jose Dolz. 2021.
\newblock Few-shot segmentation without meta-learning: A good transductive
  inference is all you need?
\newblock In \emph{Proceedings of the IEEE/CVF Conference on Computer Vision
  and Pattern Recognition}, pages 13979--13988.

\bibitem[{Boudiaf et~al.(2020)Boudiaf, Ziko, Rony, Dolz, Piantanida, and
  Ben~Ayed}]{boudiaf2020information}
Malik Boudiaf, Imtiaz Ziko, J{\'e}r{\^o}me Rony, Jos{\'e} Dolz, Pablo
  Piantanida, and Ismail Ben~Ayed. 2020.
\newblock Information maximization for few-shot learning.
\newblock \emph{Advances in Neural Information Processing Systems},
  33:2445--2457.

\bibitem[{Brown et~al.(2020)Brown, Mann, Ryder, Subbiah, Kaplan, Dhariwal,
  Neelakantan, Shyam, Sastry, Askell et~al.}]{brown2020language}
Tom Brown, Benjamin Mann, Nick Ryder, Melanie Subbiah, Jared~D Kaplan, Prafulla
  Dhariwal, Arvind Neelakantan, Pranav Shyam, Girish Sastry, Amanda Askell,
  et~al. 2020.
\newblock Language models are few-shot learners.
\newblock \emph{Advances in neural information processing systems},
  33:1877--1901.

\bibitem[{Cardoso(1997)}]{cardoso1997infomax}
J-F Cardoso. 1997.
\newblock Infomax and maximum likelihood for blind source separation.
\newblock \emph{IEEE Signal processing letters}, 4(4):112--114.

\bibitem[{Casanueva et~al.(2020)Casanueva, Tem{\v{c}}inas, Gerz, Henderson, and
  Vuli{\'c}}]{casanueva-etal-2020-efficient}
I{\~n}igo Casanueva, Tadas Tem{\v{c}}inas, Daniela Gerz, Matthew Henderson, and
  Ivan Vuli{\'c}. 2020.
\newblock Efficient intent detection with dual sentence encoders.
\newblock In \emph{Proceedings of the 2nd Workshop on Natural Language
  Processing for Conversational AI}.

\bibitem[{Chapuis et~al.(2020)Chapuis, Colombo, Manica, Labeau, and
  Clavel}]{chapuis-etal-2020-hierarchical}
Emile Chapuis, Pierre Colombo, Matteo Manica, Matthieu Labeau, and Chlo{\'e}
  Clavel. 2020.
\newblock \href {https://doi.org/10.18653/v1/2020.findings-emnlp.239}
  {Hierarchical pre-training for sequence labelling in spoken dialog}.
\newblock In \emph{Findings of the Association for Computational Linguistics:
  EMNLP 2020}, pages 2636--2648, Online. Association for Computational
  Linguistics.

\bibitem[{Chen et~al.(2021{\natexlab{a}})Chen, Tworek, Jun, Yuan, Pinto,
  Kaplan, Edwards, Burda, Joseph, Brockman et~al.}]{chen2021evaluating}
Mark Chen, Jerry Tworek, Heewoo Jun, Qiming Yuan, Henrique Ponde de~Oliveira
  Pinto, Jared Kaplan, Harri Edwards, Yuri Burda, Nicholas Joseph, Greg
  Brockman, et~al. 2021{\natexlab{a}}.
\newblock Evaluating large language models trained on code.
\newblock \emph{arXiv preprint arXiv:2107.03374}.

\bibitem[{Chen et~al.(2021{\natexlab{b}})Chen, Zhong, Zha, Karypis, and
  He}]{chen2021meta}
Yanda Chen, Ruiqi Zhong, Sheng Zha, George Karypis, and He~He.
  2021{\natexlab{b}}.
\newblock Meta-learning via language model in-context tuning.
\newblock \emph{arXiv preprint arXiv:2110.07814}.

\bibitem[{Chowdhery et~al.(2022)Chowdhery, Narang, Devlin, Bosma, Mishra,
  Roberts, Barham, Chung, Sutton, Gehrmann, Schuh, Shi, Tsvyashchenko, Maynez,
  Rao, Barnes, Tay, Shazeer, Prabhakaran, Reif, Du, Hutchinson, Pope, Bradbury,
  Austin, Isard, Gur-Ari, Yin, Duke, Levskaya, Ghemawat, Dev, Michalewski,
  Garcia, Misra, Robinson, Fedus, Zhou, Ippolito, Luan, Lim, Zoph, Spiridonov,
  Sepassi, Dohan, Agrawal, Omernick, Dai, Pillai, Pellat, Lewkowycz, Moreira,
  Child, Polozov, Lee, Zhou, Wang, Saeta, Diaz, Firat, Catasta, Wei,
  Meier-Hellstern, Eck, Dean, Petrov, and Fiedel}]{chowdhery2022palm}
Aakanksha Chowdhery, Sharan Narang, Jacob Devlin, Maarten Bosma, Gaurav Mishra,
  Adam Roberts, Paul Barham, Hyung~Won Chung, Charles Sutton, Sebastian
  Gehrmann, Parker Schuh, Kensen Shi, Sasha Tsvyashchenko, Joshua Maynez,
  Abhishek Rao, Parker Barnes, Yi~Tay, Noam Shazeer, Vinodkumar Prabhakaran,
  Emily Reif, Nan Du, Ben Hutchinson, Reiner Pope, James Bradbury, Jacob
  Austin, Michael Isard, Guy Gur-Ari, Pengcheng Yin, Toju Duke, Anselm
  Levskaya, Sanjay Ghemawat, Sunipa Dev, Henryk Michalewski, Xavier Garcia,
  Vedant Misra, Kevin Robinson, Liam Fedus, Denny Zhou, Daphne Ippolito, David
  Luan, Hyeontaek Lim, Barret Zoph, Alexander Spiridonov, Ryan Sepassi, David
  Dohan, Shivani Agrawal, Mark Omernick, Andrew~M. Dai,
  Thanumalayan~Sankaranarayana Pillai, Marie Pellat, Aitor Lewkowycz, Erica
  Moreira, Rewon Child, Oleksandr Polozov, Katherine Lee, Zongwei Zhou, Xuezhi
  Wang, Brennan Saeta, Mark Diaz, Orhan Firat, Michele Catasta, Jason Wei,
  Kathy Meier-Hellstern, Douglas Eck, Jeff Dean, Slav Petrov, and Noah Fiedel.
  2022.
\newblock \href {http://arxiv.org/abs/2204.02311} {Palm: Scaling language
  modeling with pathways}.

\bibitem[{Colombo et~al.(2021{\natexlab{a}})Colombo, Chapuis, Labeau, and
  Clavel}]{colombo-etal-2021-code}
Pierre Colombo, Emile Chapuis, Matthieu Labeau, and Chlo{\'e} Clavel.
  2021{\natexlab{a}}.
\newblock \href {https://doi.org/10.18653/v1/2021.emnlp-main.656}
  {Code-switched inspired losses for spoken dialog representations}.
\newblock In \emph{Proceedings of the 2021 Conference on Empirical Methods in
  Natural Language Processing}, pages 8320--8337, Online and Punta Cana,
  Dominican Republic. Association for Computational Linguistics.

\bibitem[{Colombo et~al.(2021{\natexlab{b}})Colombo, Chapuis, Labeau, and
  Clavel}]{colombo-etal-2021-improving}
Pierre Colombo, Emile Chapuis, Matthieu Labeau, and Chlo{\'e} Clavel.
  2021{\natexlab{b}}.
\newblock \href {https://doi.org/10.18653/v1/2021.emnlp-main.21} {Improving
  multimodal fusion via mutual dependency maximisation}.
\newblock In \emph{Proceedings of the 2021 Conference on Empirical Methods in
  Natural Language Processing}, pages 231--245, Online and Punta Cana,
  Dominican Republic. Association for Computational Linguistics.

\bibitem[{Colombo et~al.(2021{\natexlab{c}})Colombo, Clavel, Yack, and
  Varni}]{colombo-etal-2021-beam}
Pierre Colombo, Chlo{\'e} Clavel, Chouchang Yack, and Giovanna Varni.
  2021{\natexlab{c}}.
\newblock \href {https://aclanthology.org/2021.icnlsp-1.16} {Beam search with
  bidirectional strategies for neural response generation}.
\newblock In \emph{Proceedings of the 4th International Conference on Natural
  Language and Speech Processing (ICNLSP 2021)}, pages 139--146, Trento, Italy.
  Association for Computational Linguistics.

\bibitem[{Colombo et~al.(2022{\natexlab{a}})Colombo, Dadalto, Staerman, Noiry,
  and Piantanida}]{colombo2022beyond}
Pierre Colombo, Eduardo Dadalto, Guillaume Staerman, Nathan Noiry, and Pablo
  Piantanida. 2022{\natexlab{a}}.
\newblock Beyond mahalanobis distance for textual ood detection.
\newblock \emph{Advances in Neural Information Processing Systems},
  35:17744--17759.

\bibitem[{Colombo et~al.(2022{\natexlab{b}})Colombo, Noiry, Irurozki, and
  Cl{\'e}men{\c{c}}on}]{colombo2022best}
Pierre Colombo, Nathan Noiry, Ekhine Irurozki, and St{\'e}phan
  Cl{\'e}men{\c{c}}on. 2022{\natexlab{b}}.
\newblock What are the best systems? new perspectives on nlp benchmarking.
\newblock \emph{Advances in Neural Information Processing Systems},
  35:26915--26932.

\bibitem[{Colombo et~al.(2022{\natexlab{c}})Colombo, Peyrard, Noiry, West, and
  Piantanida}]{colombo2022glass}
Pierre Colombo, Maxime Peyrard, Nathan Noiry, Robert West, and Pablo
  Piantanida. 2022{\natexlab{c}}.
\newblock The glass ceiling of automatic evaluation in natural language
  generation.
\newblock \emph{arXiv preprint arXiv:2208.14585}.

\bibitem[{Colombo et~al.(2021{\natexlab{d}})Colombo, Piantanida, and
  Clavel}]{colombo-etal-2021-novel}
Pierre Colombo, Pablo Piantanida, and Chlo{\'e} Clavel. 2021{\natexlab{d}}.
\newblock \href {https://doi.org/10.18653/v1/2021.acl-long.511} {A novel
  estimator of mutual information for learning to disentangle textual
  representations}.
\newblock In \emph{Proceedings of the 59th Annual Meeting of the Association
  for Computational Linguistics and the 11th International Joint Conference on
  Natural Language Processing (Volume 1: Long Papers)}, pages 6539--6550,
  Online. Association for Computational Linguistics.

\bibitem[{Colombo et~al.(2021{\natexlab{e}})Colombo, Staerman, Clavel, and
  Piantanida}]{colombo-etal-2021-automatic}
Pierre Colombo, Guillaume Staerman, Chlo{\'e} Clavel, and Pablo Piantanida.
  2021{\natexlab{e}}.
\newblock \href {https://doi.org/10.18653/v1/2021.emnlp-main.817} {Automatic
  text evaluation through the lens of {W}asserstein barycenters}.
\newblock In \emph{Proceedings of the 2021 Conference on Empirical Methods in
  Natural Language Processing}, pages 10450--10466, Online and Punta Cana,
  Dominican Republic. Association for Computational Linguistics.

\bibitem[{Colombo et~al.(2022{\natexlab{d}})Colombo, Staerman, Noiry, and
  Piantanida}]{colombo-etal-2022-learning}
Pierre Colombo, Guillaume Staerman, Nathan Noiry, and Pablo Piantanida.
  2022{\natexlab{d}}.
\newblock \href {https://doi.org/10.18653/v1/2022.acl-long.187} {Learning
  disentangled textual representations via statistical measures of similarity}.
\newblock In \emph{Proceedings of the 60th Annual Meeting of the Association
  for Computational Linguistics (Volume 1: Long Papers)}, pages 2614--2630,
  Dublin, Ireland. Association for Computational Linguistics.

\bibitem[{Colombo et~al.(2019)Colombo, Witon, Modi, Kennedy, and
  Kapadia}]{colombo-etal-2019-affect}
Pierre Colombo, Wojciech Witon, Ashutosh Modi, James Kennedy, and Mubbasir
  Kapadia. 2019.
\newblock \href {https://doi.org/10.18653/v1/N19-1374} {Affect-driven dialog
  generation}.
\newblock In \emph{Proceedings of the 2019 Conference of the North {A}merican
  Chapter of the Association for Computational Linguistics: Human Language
  Technologies, Volume 1 (Long and Short Papers)}, pages 3734--3743,
  Minneapolis, Minnesota. Association for Computational Linguistics.

\bibitem[{Commission(2016)}]{eu2016gdpr}
European Commission. 2016.
\newblock \href {https://eur-lex.europa.eu/eli/reg/2016/679/oj} {Regulation
  (eu) 2016/679 of the european parliament and of the council of 27 april 2016
  on the protection of natural persons with regard to the processing of
  personal data and on the free movement of such data, and repealing directive
  95/46/ec (general data protection regulation)}.

\bibitem[{Commission(2020)}]{eu2020datagovernance}
European Commission. 2020.
\newblock \href
  {https://eur-lex.europa.eu/legal-content/EN/TXT/PDF/?uri=CELEX:52020PC0767&from=EN}
  {Proposal for a regulation of the european parliament and of the council on
  european data governance (data governance act)}.
\newblock COM(2020) 767 final.

\bibitem[{Conneau et~al.(2019)Conneau, Khandelwal, Goyal, Chaudhary, Wenzek,
  Guzm{\'a}n, Grave, Ott, Zettlemoyer, and Stoyanov}]{conneau2019unsupervised}
Alexis Conneau, Kartikay Khandelwal, Naman Goyal, Vishrav Chaudhary, Guillaume
  Wenzek, Francisco Guzm{\'a}n, Edouard Grave, Myle Ott, Luke Zettlemoyer, and
  Veselin Stoyanov. 2019.
\newblock Unsupervised cross-lingual representation learning at scale.
\newblock \emph{arXiv preprint arXiv:1911.02116}.

\bibitem[{Conneau et~al.(2020)Conneau, Khandelwal, Goyal, Chaudhary, Wenzek,
  Guzm{\'a}n, Grave, Ott, Zettlemoyer, and
  Stoyanov}]{conneau-etal-2020-unsupervised}
Alexis Conneau, Kartikay Khandelwal, Naman Goyal, Vishrav Chaudhary, Guillaume
  Wenzek, Francisco Guzm{\'a}n, Edouard Grave, Myle Ott, Luke Zettlemoyer, and
  Veselin Stoyanov. 2020.
\newblock Unsupervised cross-lingual representation learning at scale.
\newblock In \emph{Proceedings of the 58th Annual Meeting of the Association
  for Computational Linguistics}.

\bibitem[{Cover(1999)}]{cover1999elements}
Thomas~M Cover. 1999.
\newblock \emph{Elements of information theory}.
\newblock John Wiley \& Sons.

\bibitem[{Cui et~al.(2022)Cui, Hu, Ding, Huang, and Liu}]{cui2022prototypical}
Ganqu Cui, Shengding Hu, Ning Ding, Longtao Huang, and Zhiyuan Liu. 2022.
\newblock Prototypical verbalizer for prompt-based few-shot tuning.
\newblock \emph{arXiv preprint arXiv:2203.09770}.

\bibitem[{Darrin et~al.(2022)Darrin, Piantanida, and
  Colombo}]{darrin2022rainproof}
Maxime Darrin, Pablo Piantanida, and Pierre Colombo. 2022.
\newblock Rainproof: An umbrella to shield text generators from
  out-of-distribution data.
\newblock \emph{arXiv preprint arXiv:2212.09171}.

\bibitem[{Darrin et~al.(2023)Darrin, Staerman, Gomes, Cheung, Piantanida, and
  Colombo}]{darrin2023unsupervised}
Maxime Darrin, Guillaume Staerman, Eduardo Dadalto~C{\^a}mara Gomes, Jackie~CK
  Cheung, Pablo Piantanida, and Pierre Colombo. 2023.
\newblock Unsupervised layer-wise score aggregation for textual ood detection.
\newblock \emph{arXiv preprint arXiv:2302.09852}.

\bibitem[{Demszky et~al.(2020)Demszky, Movshovitz-Attias, Ko, Cowen, Nemade,
  and Ravi}]{demszky2020goemotions}
Dorottya Demszky, Dana Movshovitz-Attias, Jeongwoo Ko, Alan Cowen, Gaurav
  Nemade, and Sujith Ravi. 2020.
\newblock {GoEmotions: A Dataset of Fine-Grained Emotions}.
\newblock In \emph{58th Annual Meeting of the Association for Computational
  Linguistics (ACL)}.

\bibitem[{Devlin et~al.(2018)Devlin, Chang, Lee, and
  Toutanova}]{devlin2018bert}
Jacob Devlin, Ming-Wei Chang, Kenton Lee, and Kristina Toutanova. 2018.
\newblock Bert: Pre-training of deep bidirectional transformers for language
  understanding.
\newblock \emph{arXiv preprint arXiv:1810.04805}.

\bibitem[{Dhillon et~al.(2019)Dhillon, Chaudhari, Ravichandran, and
  Soatto}]{dhillon2019baseline}
Guneet~S Dhillon, Pratik Chaudhari, Avinash Ravichandran, and Stefano Soatto.
  2019.
\newblock A baseline for few-shot image classification.
\newblock \emph{arXiv preprint arXiv:1909.02729}.

\bibitem[{Ding et~al.(2021)Ding, Hu, Zhao, Chen, Liu, Zheng, and
  Sun}]{ding2021openprompt}
Ning Ding, Shengding Hu, Weilin Zhao, Yulin Chen, Zhiyuan Liu, Hai-Tao Zheng,
  and Maosong Sun. 2021.
\newblock Openprompt: An open-source framework for prompt-learning.
\newblock \emph{arXiv preprint arXiv:2111.01998}.

\bibitem[{Fan et~al.(2021)Fan, Bhosale, Schwenk, Ma, El-Kishky, Goyal, Baines,
  Celebi, Wenzek, Chaudhary et~al.}]{fan2021beyond}
Angela Fan, Shruti Bhosale, Holger Schwenk, Zhiyi Ma, Ahmed El-Kishky,
  Siddharth Goyal, Mandeep Baines, Onur Celebi, Guillaume Wenzek, Vishrav
  Chaudhary, et~al. 2021.
\newblock Beyond english-centric multilingual machine translation.
\newblock \emph{The Journal of Machine Learning Research}, 22(1):4839--4886.

\bibitem[{Finn et~al.(2017{\natexlab{a}})Finn, Abbeel, and
  Levine}]{finn2017model}
Chelsea Finn, Pieter Abbeel, and Sergey Levine. 2017{\natexlab{a}}.
\newblock Model-agnostic meta-learning for fast adaptation of deep networks.
\newblock In \emph{International conference on machine learning}, pages
  1126--1135. PMLR.

\bibitem[{Finn et~al.(2017{\natexlab{b}})Finn, Abbeel, and
  Levine}]{finn-17-maml}
Chelsea Finn, Pieter Abbeel, and Sergey Levine. 2017{\natexlab{b}}.
\newblock Model-agnostic meta-learning for fast adaptation of deep networks.
\newblock In \emph{Proceedings of the 34th International Conference on Machine
  Learning}.

\bibitem[{Gao et~al.(2020)Gao, Fisch, and Chen}]{gao2020making}
Tianyu Gao, Adam Fisch, and Danqi Chen. 2020.
\newblock Making pre-trained language models better few-shot learners.
\newblock \emph{arXiv preprint arXiv:2012.15723}.

\bibitem[{Gao et~al.(2019)Gao, Han, Liu, and Sun}]{gao2019hybrid}
Tianyu Gao, Xu~Han, Zhiyuan Liu, and Maosong Sun. 2019.
\newblock Hybrid attention-based prototypical networks for noisy few-shot
  relation classification.
\newblock In \emph{Proceedings of the AAAI conference on artificial
  intelligence}, volume~33, pages 6407--6414.

\bibitem[{Grandvalet and Bengio(2004)}]{grandvalet2004semi}
Yves Grandvalet and Yoshua Bengio. 2004.
\newblock Semi-supervised learning by entropy minimization.
\newblock \emph{Advances in neural information processing systems}, 17.

\bibitem[{Himmi et~al.(2023)Himmi, Irurozki, Noiry, Clemencon, and
  Colombo}]{himmi2023towards}
Anas Himmi, Ekhine Irurozki, Nathan Noiry, Stephan Clemencon, and Pierre
  Colombo. 2023.
\newblock Towards more robust nlp system evaluation: Handling missing scores in
  benchmarks.
\newblock \emph{arXiv preprint arXiv:2305.10284}.

\bibitem[{Hinton et~al.(2015)Hinton, Vinyals, and
  Dean}]{Hinton2015DistillingTK}
Geoffrey~E. Hinton, Oriol Vinyals, and Jeffrey Dean. 2015.
\newblock Distilling the knowledge in a neural network.
\newblock \emph{ArXiv}.

\bibitem[{Hospedales et~al.(2021)Hospedales, Antoniou, Micaelli, and
  Storkey}]{hospedales2021meta}
Timothy Hospedales, Antreas Antoniou, Paul Micaelli, and Amos Storkey. 2021.
\newblock Meta-learning in neural networks: A survey.
\newblock \emph{IEEE transactions on pattern analysis and machine
  intelligence}, 44(9):5149--5169.

\bibitem[{Houlsby et~al.(2019)Houlsby, Giurgiu, Jastrzebski, Morrone,
  De~Laroussilhe, Gesmundo, Attariyan, and Gelly}]{houlsby2019parameter}
Neil Houlsby, Andrei Giurgiu, Stanislaw Jastrzebski, Bruna Morrone, Quentin
  De~Laroussilhe, Andrea Gesmundo, Mona Attariyan, and Sylvain Gelly. 2019.
\newblock Parameter-efficient transfer learning for nlp.
\newblock In \emph{International Conference on Machine Learning}, pages
  2790--2799. PMLR.

\bibitem[{Hu et~al.(2021{\natexlab{a}})Hu, Ding, Wang, Liu, Li, and
  Sun}]{hu2021knowledgeable}
Shengding Hu, Ning Ding, Huadong Wang, Zhiyuan Liu, Juanzi Li, and Maosong Sun.
  2021{\natexlab{a}}.
\newblock Knowledgeable prompt-tuning: Incorporating knowledge into prompt
  verbalizer for text classification.
\newblock \emph{arXiv preprint arXiv:2108.02035}.

\bibitem[{Hu et~al.(2017)Hu, Miyato, Tokui, Matsumoto, and
  Sugiyama}]{Hu2017-imsat}
Weihua Hu, Takeru Miyato, Seiya Tokui, Eiichi Matsumoto, and Masashi Sugiyama.
  2017.
\newblock Learning discrete representations via information maximizing
  self-augmented training.
\newblock In \emph{Proceedings of the 34th International Conference on Machine
  Learning - Volume 70}.

\bibitem[{Hu et~al.(2021{\natexlab{b}})Hu, Gripon, and
  Pateux}]{hu2021leveraging}
Yuqing Hu, Vincent Gripon, and St{\'e}phane Pateux. 2021{\natexlab{b}}.
\newblock Leveraging the feature distribution in transfer-based few-shot
  learning.
\newblock In \emph{Artificial Neural Networks and Machine Learning--ICANN 2021:
  30th International Conference on Artificial Neural Networks, Bratislava,
  Slovakia, September 14--17, 2021, Proceedings, Part II 30}, pages 487--499.
  Springer.

\bibitem[{Kaplan et~al.(2020)Kaplan, McCandlish, Henighan, Brown, Chess, Child,
  Gray, Radford, Wu, and Amodei}]{kaplan2020scaling}
Jared Kaplan, Sam McCandlish, Tom Henighan, Tom~B Brown, Benjamin Chess, Rewon
  Child, Scott Gray, Alec Radford, Jeffrey Wu, and Dario Amodei. 2020.
\newblock Scaling laws for neural language models.
\newblock \emph{arXiv preprint arXiv:2001.08361}.

\bibitem[{Karimi~Mahabadi et~al.(2022)Karimi~Mahabadi, Zettlemoyer, Henderson,
  Saeidi, Mathias, Stoyano, and Yazdani}]{karimi2022perfect}
Rabeeh Karimi~Mahabadi, Luke Zettlemoyer, James Henderson, Marzieh Saeidi,
  Lambert Mathias, Veselin Stoyano, and Majid Yazdani. 2022.
\newblock Perfect: Prompt-free and efficient few-shot learning with language
  models.
\newblock In \emph{Annual Meeting of the Association for Computational
  Linguistics}.

\bibitem[{Keung et~al.(2020)Keung, Lu, Szarvas, and
  Smith}]{keung-etal-2020-multilingual}
Phillip Keung, Yichao Lu, Gy{\"o}rgy Szarvas, and Noah~A. Smith. 2020.
\newblock The multilingual {A}mazon reviews corpus.
\newblock In \emph{Proceedings of the 2020 Conference on Empirical Methods in
  Natural Language Processing (EMNLP)}.

\bibitem[{Kocoń et~al.(2023)Kocoń, Cichecki, Kaszyca, Kochanek, Szydło,
  Baran, Bielaniewicz, Gruza, Janz, Kanclerz, Kocoń, Koptyra,
  Mieleszczenko-Kowszewicz, Miłkowski, Oleksy, Piasecki, Łukasz Radliński,
  Wojtasik, Woźniak, and Kazienko}]{kocoń2023chatgpt}
Jan Kocoń, Igor Cichecki, Oliwier Kaszyca, Mateusz Kochanek, Dominika Szydło,
  Joanna Baran, Julita Bielaniewicz, Marcin Gruza, Arkadiusz Janz, Kamil
  Kanclerz, Anna Kocoń, Bartłomiej Koptyra, Wiktoria
  Mieleszczenko-Kowszewicz, Piotr Miłkowski, Marcin Oleksy, Maciej Piasecki,
  Łukasz Radliński, Konrad Wojtasik, Stanisław Woźniak, and Przemysław
  Kazienko. 2023.
\newblock \href {http://arxiv.org/abs/2302.10724} {Chatgpt: Jack of all trades,
  master of none}.

\bibitem[{Lan et~al.(2019)Lan, Chen, Goodman, Gimpel, Sharma, and
  Soricut}]{lan2019albert}
Zhenzhong Lan, Mingda Chen, Sebastian Goodman, Kevin Gimpel, Piyush Sharma, and
  Radu Soricut. 2019.
\newblock Albert: A lite bert for self-supervised learning of language
  representations.
\newblock \emph{arXiv preprint arXiv:1909.11942}.

\bibitem[{Larson et~al.(2019)Larson, Mahendran, Peper, Clarke, Lee, Hill,
  Kummerfeld, Leach, Laurenzano, Tang, and Mars}]{larson-etal-2019-evaluation}
Stefan Larson, Anish Mahendran, Joseph~J. Peper, Christopher Clarke, Andrew
  Lee, Parker Hill, Jonathan~K. Kummerfeld, Kevin Leach, Michael~A. Laurenzano,
  Lingjia Tang, and Jason Mars. 2019.
\newblock \href {https://www.aclweb.org/anthology/D19-1131} {An evaluation
  dataset for intent classification and out-of-scope prediction}.
\newblock In \emph{Proceedings of the 2019 Conference on Empirical Methods in
  Natural Language Processing and the 9th International Joint Conference on
  Natural Language Processing (EMNLP-IJCNLP)}.

\bibitem[{Lee et~al.(2019)Lee, Maji, Ravichandran, and Soatto}]{lee2019meta}
Kwonjoon Lee, Subhransu Maji, Avinash Ravichandran, and Stefano Soatto. 2019.
\newblock Meta-learning with differentiable convex optimization.
\newblock In \emph{Proceedings of the IEEE/CVF conference on computer vision
  and pattern recognition}, pages 10657--10665.

\bibitem[{Lehman et~al.(2023)Lehman, Hernandez, Mahajan, Wulff, Smith, Ziegler,
  Nadler, Szolovits, Johnson, and Alsentzer}]{lehman2023need}
Eric Lehman, Evan Hernandez, Diwakar Mahajan, Jonas Wulff, Micah~J. Smith,
  Zachary Ziegler, Daniel Nadler, Peter Szolovits, Alistair Johnson, and Emily
  Alsentzer. 2023.
\newblock \href {http://arxiv.org/abs/2302.08091} {Do we still need clinical
  language models?}

\bibitem[{Lester et~al.(2021)Lester, Al-Rfou, and Constant}]{lester2021power}
Brian Lester, Rami Al-Rfou, and Noah Constant. 2021.
\newblock The power of scale for parameter-efficient prompt tuning.
\newblock \emph{arXiv preprint arXiv:2104.08691}.

\bibitem[{Lhoest et~al.(2021)Lhoest, del Moral, Jernite, Thakur, von Platen,
  Patil, Chaumond, Drame, Plu, Tunstall et~al.}]{lhoest2021datasets}
Quentin Lhoest, Albert~Villanova del Moral, Yacine Jernite, Abhishek Thakur,
  Patrick von Platen, Suraj Patil, Julien Chaumond, Mariama Drame, Julien Plu,
  Lewis Tunstall, et~al. 2021.
\newblock Datasets: A community library for natural language processing.
\newblock \emph{arXiv preprint arXiv:2109.02846}.

\bibitem[{Lichtenstein et~al.(2020)Lichtenstein, Sattigeri, Feris, Giryes, and
  Karlinsky}]{lichtenstein2020tafssl}
Moshe Lichtenstein, Prasanna Sattigeri, Rogerio Feris, Raja Giryes, and Leonid
  Karlinsky. 2020.
\newblock Tafssl: Task-adaptive feature sub-space learning for few-shot
  classification.
\newblock In \emph{Computer Vision--ECCV 2020: 16th European Conference,
  Glasgow, UK, August 23--28, 2020, Proceedings, Part VII}, pages 522--539.
  Springer.

\bibitem[{Linsker(1988)}]{linsker1988self}
Ralph Linsker. 1988.
\newblock Self-organization in a perceptual network.
\newblock \emph{Computer}, 21(3):105--117.

\bibitem[{Liu et~al.(2022)Liu, Tam, Muqeeth, Mohta, Huang, Bansal, and
  Raffel}]{liu2022few}
Haokun Liu, Derek Tam, Mohammed Muqeeth, Jay Mohta, Tenghao Huang, Mohit
  Bansal, and Colin Raffel. 2022.
\newblock Few-shot parameter-efficient fine-tuning is better and cheaper than
  in-context learning.
\newblock \emph{arXiv preprint arXiv:2205.05638}.

\bibitem[{Liu et~al.(2023)Liu, Yuan, Fu, Jiang, Hayashi, and
  Neubig}]{liu2023pre}
Pengfei Liu, Weizhe Yuan, Jinlan Fu, Zhengbao Jiang, Hiroaki Hayashi, and
  Graham Neubig. 2023.
\newblock Pre-train, prompt, and predict: A systematic survey of prompting
  methods in natural language processing.
\newblock \emph{ACM Computing Surveys}, 55(9):1--35.

\bibitem[{Liu et~al.(2019{\natexlab{a}})Liu, Lee, Park, Kim, Yang, Hwang, and
  Yang}]{liu2018learning}
Yanbin Liu, Juho Lee, Minseop Park, Saehoon Kim, Eunho Yang, Sung~Ju Hwang, and
  Yi~Yang. 2019{\natexlab{a}}.
\newblock Learning to propagate labels: Transductive propagation network for
  few-shot learning.
\newblock \emph{ICLR}.

\bibitem[{Liu et~al.(2019{\natexlab{b}})Liu, Ott, Goyal, Du, Joshi, Chen, Levy,
  Lewis, Zettlemoyer, and Stoyanov}]{liu2019roberta}
Yinhan Liu, Myle Ott, Naman Goyal, Jingfei Du, Mandar Joshi, Danqi Chen, Omer
  Levy, Mike Lewis, Luke Zettlemoyer, and Veselin Stoyanov. 2019{\natexlab{b}}.
\newblock Roberta: A robustly optimized bert pretraining approach.
\newblock \emph{arXiv preprint arXiv:1907.11692}.

\bibitem[{Logan~IV et~al.(2021)Logan~IV, Bala{\v{z}}evi{\'c}, Wallace, Petroni,
  Singh, and Riedel}]{logan2021cutting}
Robert~L Logan~IV, Ivana Bala{\v{z}}evi{\'c}, Eric Wallace, Fabio Petroni,
  Sameer Singh, and Sebastian Riedel. 2021.
\newblock Cutting down on prompts and parameters: Simple few-shot learning with
  language models.
\newblock \emph{arXiv preprint arXiv:2106.13353}.

\bibitem[{Mahabadi et~al.(2022)Mahabadi, Zettlemoyer, Henderson, Saeidi,
  Mathias, Stoyanov, and Yazdani}]{mahabadi2022perfect}
Rabeeh~Karimi Mahabadi, Luke Zettlemoyer, James Henderson, Marzieh Saeidi,
  Lambert Mathias, Veselin Stoyanov, and Majid Yazdani. 2022.
\newblock Perfect: Prompt-free and efficient few-shot learning with language
  models.
\newblock \emph{arXiv preprint arXiv:2204.01172}.

\bibitem[{Min et~al.(2021)Min, Lewis, Zettlemoyer, and
  Hajishirzi}]{min2021metaicl}
Sewon Min, Mike Lewis, Luke Zettlemoyer, and Hannaneh Hajishirzi. 2021.
\newblock Metaicl: Learning to learn in context.
\newblock \emph{arXiv preprint arXiv:2110.15943}.

\bibitem[{Min et~al.(2022)Min, Lyu, Holtzman, Artetxe, Lewis, Hajishirzi, and
  Zettlemoyer}]{min2022rethinking}
Sewon Min, Xinxi Lyu, Ari Holtzman, Mikel Artetxe, Mike Lewis, Hannaneh
  Hajishirzi, and Luke Zettlemoyer. 2022.
\newblock Rethinking the role of demonstrations: What makes in-context learning
  work?
\newblock \emph{arXiv preprint arXiv:2202.12837}.

\bibitem[{Mishra et~al.(2017)Mishra, Rohaninejad, Chen, and
  Abbeel}]{Mishra2017ASN}
Nikhil Mishra, Mostafa Rohaninejad, Xi~Chen, and P.~Abbeel. 2017.
\newblock A simple neural attentive meta-learner.
\newblock In \emph{International Conference on Learning Representations}.

\bibitem[{Muennighoff et~al.(2022)Muennighoff, Tazi, Magne, and
  Reimers}]{muennighoff2022mteb}
Niklas Muennighoff, Nouamane Tazi, Lo{\"\i}c Magne, and Nils Reimers. 2022.
\newblock Mteb: Massive text embedding benchmark.
\newblock \emph{arXiv preprint arXiv:2210.07316}.

\bibitem[{OpenAI(2023)}]{openai2023gpt4}
OpenAI. 2023.
\newblock \href {http://arxiv.org/abs/2303.08774} {Gpt-4 technical report}.

\bibitem[{Oreshkin et~al.(2018)Oreshkin, Rodriguez, and
  Lacoste}]{oreshkin-2018-tadam}
Boris~N. Oreshkin, Pau Rodriguez, and Alexandre Lacoste. 2018.
\newblock Tadam: Task dependent adaptive metric for improved few-shot learning.
\newblock In \emph{Proceedings of the 32nd International Conference on Neural
  Information Processing Systems}, page 719–729.

\bibitem[{Perez et~al.(2021)Perez, Kiela, and Cho}]{perez2021true}
Ethan Perez, Douwe Kiela, and Kyunghyun Cho. 2021.
\newblock True few-shot learning with language models.
\newblock \emph{Advances in neural information processing systems},
  34:11054--11070.

\bibitem[{Pfeiffer et~al.(2020)Pfeiffer, R{\"u}ckl{\'e}, Poth, Kamath,
  Vuli{\'c}, Ruder, Cho, and Gurevych}]{pfeiffer2020adapterhub}
Jonas Pfeiffer, Andreas R{\"u}ckl{\'e}, Clifton Poth, Aishwarya Kamath, Ivan
  Vuli{\'c}, Sebastian Ruder, Kyunghyun Cho, and Iryna Gurevych. 2020.
\newblock Adapterhub: A framework for adapting transformers.
\newblock \emph{arXiv preprint arXiv:2007.07779}.

\bibitem[{Pichler et~al.(2022)Pichler, Colombo, Boudiaf, Koliander, and
  Piantanida}]{pichler2022differential}
Georg Pichler, Pierre Jean~A Colombo, Malik Boudiaf, G{\"u}nther Koliander, and
  Pablo Piantanida. 2022.
\newblock A differential entropy estimator for training neural networks.
\newblock In \emph{International Conference on Machine Learning}, pages
  17691--17715. PMLR.

\bibitem[{Picot et~al.(2023)Picot, Messina, Boudiaf, Labeau, Ayed, and
  Piantanida}]{9773978}
Marine Picot, Francisco Messina, Malik Boudiaf, Fabrice Labeau, Ismail~Ben
  Ayed, and Pablo Piantanida. 2023.
\newblock \href {https://doi.org/10.1109/TPAMI.2022.3174724} {Adversarial
  robustness via fisher-rao regularization}.
\newblock \emph{IEEE Transactions on Pattern Analysis and Machine
  Intelligence}, 45(3):2698--2710.

\bibitem[{Picot et~al.(2022{\natexlab{a}})Picot, Noiry, Piantanida, and
  Colombo}]{picot2022adversarial}
Marine Picot, Nathan Noiry, Pablo Piantanida, and Pierre Colombo.
  2022{\natexlab{a}}.
\newblock Adversarial attack detection under realistic constraints.

\bibitem[{Picot et~al.(2022{\natexlab{b}})Picot, Staerman, Granese, Noiry,
  Messina, Piantanida, and Colombo}]{picot2022simple}
Marine Picot, Guillaume Staerman, Federica Granese, Nathan Noiry, Francisco
  Messina, Pablo Piantanida, and Pierre Colombo. 2022{\natexlab{b}}.
\newblock A simple unsupervised data depth-based method to detect adversarial
  images.

\bibitem[{Radford et~al.(2019)Radford, Wu, Child, Luan, Amodei, and
  Sutskever}]{radford2019language}
Alec Radford, Jeff Wu, Rewon Child, David Luan, Dario Amodei, and Ilya
  Sutskever. 2019.
\newblock Language models are unsupervised multitask learners.

\bibitem[{Raffel et~al.(2020)Raffel, Shazeer, Roberts, Lee, Narang, Matena,
  Zhou, Li, and Liu}]{raffel2020exploring}
Colin Raffel, Noam Shazeer, Adam Roberts, Katherine Lee, Sharan Narang, Michael
  Matena, Yanqi Zhou, Wei Li, and Peter~J Liu. 2020.
\newblock Exploring the limits of transfer learning with a unified text-to-text
  transformer.
\newblock \emph{The Journal of Machine Learning Research}, 21(1):5485--5551.

\bibitem[{Raghu et~al.(2019)Raghu, Raghu, Bengio, and Vinyals}]{raghu2019rapid}
Aniruddh Raghu, Maithra Raghu, Samy Bengio, and Oriol Vinyals. 2019.
\newblock Rapid learning or feature reuse? towards understanding the
  effectiveness of maml.
\newblock \emph{arXiv preprint arXiv:1909.09157}.

\bibitem[{Reimers and Gurevych(2019)}]{reimers-gurevych-2019-sentence}
Nils Reimers and Iryna Gurevych. 2019.
\newblock Sentence-{BERT}: Sentence embeddings using {S}iamese {BERT}-networks.
\newblock In \emph{Proceedings of the 2019 Conference on Empirical Methods in
  Natural Language Processing and the 9th International Joint Conference on
  Natural Language Processing (EMNLP-IJCNLP)}.

\bibitem[{Rosenfeld et~al.(2019)Rosenfeld, Rosenfeld, Belinkov, and
  Shavit}]{rosenfeld2019constructive}
Jonathan~S Rosenfeld, Amir Rosenfeld, Yonatan Belinkov, and Nir Shavit. 2019.
\newblock A constructive prediction of the generalization error across scales.
\newblock \emph{arXiv preprint arXiv:1909.12673}.

\bibitem[{Rusu et~al.(2019)Rusu, Rao, Sygnowski, Vinyals, Pascanu, Osindero,
  and Hadsell}]{Rusu-2018-embedding-optimization}
Andrei Rusu, Dushyant Rao, Jakub Sygnowski, Oriol Vinyals, Razvan Pascanu,
  Simon Osindero, and Raia Hadsell. 2019.
\newblock Meta-learning with latent embedding optimization.

\bibitem[{Sain(1996)}]{sain1996nature}
Stephan~R Sain. 1996.
\newblock The nature of statistical learning theory.

\bibitem[{Schick and Sch{\"u}tze(2020{\natexlab{a}})}]{schick2020exploiting}
Timo Schick and Hinrich Sch{\"u}tze. 2020{\natexlab{a}}.
\newblock Exploiting cloze questions for few shot text classification and
  natural language inference.
\newblock \emph{arXiv preprint arXiv:2001.07676}.

\bibitem[{Schick and Sch{\"u}tze(2020{\natexlab{b}})}]{schick2020s}
Timo Schick and Hinrich Sch{\"u}tze. 2020{\natexlab{b}}.
\newblock It's not just size that matters: Small language models are also
  few-shot learners.
\newblock \emph{arXiv preprint arXiv:2009.07118}.

\bibitem[{Schick and Sch{\"u}tze(2022)}]{schick2022true}
Timo Schick and Hinrich Sch{\"u}tze. 2022.
\newblock True few-shot learning with prompts—a real-world perspective.
\newblock \emph{Transactions of the Association for Computational Linguistics},
  10:716--731.

\bibitem[{Snell et~al.(2017)Snell, Swersky, and Zemel}]{snell2017prototypical}
Jake Snell, Kevin Swersky, and Richard Zemel. 2017.
\newblock Prototypical networks for few-shot learning.
\newblock \emph{Advances in neural information processing systems}, 30.

\bibitem[{Solaiman(2023)}]{solaiman2023gradient}
Irene Solaiman. 2023.
\newblock The gradient of generative ai release: Methods and considerations.
\newblock \emph{arXiv preprint arXiv:2302.04844}.

\bibitem[{Song et~al.(2020)Song, Tan, Qin, Lu, and Liu}]{song2020mpnet}
Kaitao Song, Xu~Tan, Tao Qin, Jianfeng Lu, and Tie-Yan Liu. 2020.
\newblock Mpnet: Masked and permuted pre-training for language understanding.
\newblock \emph{Advances in Neural Information Processing Systems},
  33:16857--16867.

\bibitem[{Song et~al.(2022)Song, Wang, Cai, Mondal, and
  Sahoo}]{song2022comprehensive}
Yisheng Song, Ting Wang, Puyu Cai, Subrota~K Mondal, and Jyoti~Prakash Sahoo.
  2022.
\newblock A comprehensive survey of few-shot learning: Evolution, applications,
  challenges, and opportunities.
\newblock \emph{ACM Computing Surveys}.

\bibitem[{Strubell et~al.(2019)Strubell, Ganesh, and
  McCallum}]{strubell2019energy}
Emma Strubell, Ananya Ganesh, and Andrew McCallum. 2019.
\newblock \href {http://arxiv.org/abs/1906.02243} {Energy and policy
  considerations for deep learning in nlp}.

\bibitem[{Sun et~al.(2019{\natexlab{a}})Sun, Liu, Chua, and
  Schiele}]{sun2019meta}
Qianru Sun, Yaoyao Liu, Tat-Seng Chua, and Bernt Schiele. 2019{\natexlab{a}}.
\newblock Meta-transfer learning for few-shot learning.
\newblock In \emph{Proceedings of the IEEE/CVF Conference on Computer Vision
  and Pattern Recognition}, pages 403--412.

\bibitem[{Sun et~al.(2019{\natexlab{b}})Sun, Sun, Zhou, and
  Lv}]{sun2019hierarchical}
Shengli Sun, Qingfeng Sun, Kevin Zhou, and Tengchao Lv. 2019{\natexlab{b}}.
\newblock Hierarchical attention prototypical networks for few-shot text
  classification.
\newblock In \emph{Proceedings of the 2019 conference on empirical methods in
  natural language processing and the 9th international joint conference on
  natural language processing (EMNLP-IJCNLP)}, pages 476--485.

\bibitem[{Sung et~al.(2018{\natexlab{a}})Sung, Yang, Zhang, Xiang, Torr, and
  Hospedales}]{sung-2018-compare}
Flood Sung, Yongxin Yang, Li~Zhang, Tao Xiang, Philip Torr, and Timothy
  Hospedales. 2018{\natexlab{a}}.
\newblock \href {https://doi.org/10.1109/CVPR.2018.00131} {Learning to compare:
  Relation network for few-shot learning}.

\bibitem[{Sung et~al.(2018{\natexlab{b}})Sung, Yang, Zhang, Xiang, Torr, and
  Hospedales}]{sung2018learning}
Flood Sung, Yongxin Yang, Li~Zhang, Tao Xiang, Philip~HS Torr, and Timothy~M
  Hospedales. 2018{\natexlab{b}}.
\newblock Learning to compare: Relation network for few-shot learning.
\newblock In \emph{Proceedings of the IEEE conference on computer vision and
  pattern recognition}, pages 1199--1208.

\bibitem[{Tam et~al.(2021)Tam, Menon, Bansal, Srivastava, and
  Raffel}]{tam2021improving}
Derek Tam, Rakesh~R Menon, Mohit Bansal, Shashank Srivastava, and Colin Raffel.
  2021.
\newblock Improving and simplifying pattern exploiting training.
\newblock \emph{arXiv preprint arXiv:2103.11955}.

\bibitem[{Taylor(1953)}]{taylor1953cloze}
Wilson~L Taylor. 1953.
\newblock “cloze procedure”: A new tool for measuring readability.
\newblock \emph{Journalism quarterly}, 30(4):415--433.

\bibitem[{Touvron et~al.(2023)Touvron, Lavril, Izacard, Martinet, Lachaux,
  Lacroix, Rozi{\`e}re, Goyal, Hambro, Azhar et~al.}]{touvron2023llama}
Hugo Touvron, Thibaut Lavril, Gautier Izacard, Xavier Martinet, Marie-Anne
  Lachaux, Timoth{\'e}e Lacroix, Baptiste Rozi{\`e}re, Naman Goyal, Eric
  Hambro, Faisal Azhar, et~al. 2023.
\newblock Llama: Open and efficient foundation language models.
\newblock \emph{arXiv preprint arXiv:2302.13971}.

\bibitem[{Vapnik(1999)}]{vapnik1999overview}
Vladimir~N Vapnik. 1999.
\newblock An overview of statistical learning theory.
\newblock \emph{IEEE transactions on neural networks}, 10(5):988--999.

\bibitem[{Vinyals et~al.(2016)Vinyals, Blundell, Lillicrap, kavukcuoglu, and
  Wierstra}]{vinyals-2016-matching}
Oriol Vinyals, Charles Blundell, Timothy Lillicrap, koray kavukcuoglu, and Daan
  Wierstra. 2016.
\newblock Matching networks for one shot learning.
\newblock In \emph{Advances in Neural Information Processing Systems}.

\bibitem[{Wang et~al.(2020)Wang, Wei, Dong, Bao, Yang, and
  Zhou}]{wang2020minilm}
Wenhui Wang, Furu Wei, Li~Dong, Hangbo Bao, Nan Yang, and Ming Zhou. 2020.
\newblock Minilm: Deep self-attention distillation for task-agnostic
  compression of pre-trained transformers.
\newblock \emph{Advances in Neural Information Processing Systems},
  33:5776--5788.

\bibitem[{Wei et~al.(2022)Wei, Wang, Schuurmans, Bosma, Chi, Le, and
  Zhou}]{wei2022chain}
Jason Wei, Xuezhi Wang, Dale Schuurmans, Maarten Bosma, Ed~Chi, Quoc Le, and
  Denny Zhou. 2022.
\newblock Chain of thought prompting elicits reasoning in large language
  models.
\newblock \emph{arXiv preprint arXiv:2201.11903}.

\bibitem[{Yang et~al.(2019)Yang, Dai, Yang, Carbonell, Salakhutdinov, and
  Le}]{yang2019xlnet}
Zhilin Yang, Zihang Dai, Yiming Yang, Jaime Carbonell, Russ~R Salakhutdinov,
  and Quoc~V Le. 2019.
\newblock Xlnet: Generalized autoregressive pretraining for language
  understanding.
\newblock \emph{Advances in neural information processing systems}, 32.

\bibitem[{Ye et~al.(2021)Ye, Lin, and Ren}]{ye2021crossfit}
Qinyuan Ye, Bill~Yuchen Lin, and Xiang Ren. 2021.
\newblock Crossfit: A few-shot learning challenge for cross-task generalization
  in nlp.
\newblock \emph{arXiv preprint arXiv:2104.08835}.

\bibitem[{Ziko et~al.(2020)Ziko, Dolz, Granger, and Ayed}]{ziko2020laplacian}
Imtiaz Ziko, Jose Dolz, Eric Granger, and Ismail~Ben Ayed. 2020.
\newblock Laplacian regularized few-shot learning.
\newblock In \emph{International conference on machine learning}, pages
  11660--11670. PMLR.

\end{thebibliography}
\bibliographystyle{acl_natbib}
\newpage
\clearpage
\appendix
\onecolumn

\startcontents[sections]
\printcontents[sections]{l}{1}{\setcounter{tocdepth}{2}}

\section{Proof of Proposition~\ref{th:one}}\label{sec:proofs}

In this Appendix, we prove the inequality (\autoref{eq:kl_bound1}) provided in Proposition~\ref{th:one}.  The right-hand side of (\autoref{eq:kl_bound1}) follows straightforwardly from the definition of $R_Q^{I}( \alpha)$ and the non-negativity of the Shannon entropy. In order to prove the first inequality, we need to introduce the following intermediate result. 

For any arbitrary random variable (r.v) $X$ and countable r.v $Y$, and any real number $\beta$, let 
$$
I_\beta (X; Y ) := -\mathbb{E}_{X^\star Y} \log  \mathbb{E}_{X} \left[\frac{P(Y|X)}{P(Y|X^\star)} \right]^\beta,  
$$
where the r.v $X^\star$ follows the same distribution than $X$.  Notice that it is obvious that $I_1 (X; Y ) = I (X; Y )$, where $I (X; Y )$ is Shannon Mutual Information. 

\begin{lemma}\label{lemma-inequality}
For any arbitrary r.v. $X$ and countable r.v. $Y$, we have 
$$
I (X; Y ) \geq I_\beta (X; Y ), \   \textrm{ for $\, 0\leq \beta \leq 1$. }
$$
\end{lemma}
\textit{Proof of the lemma:} We must show that the different of $I (X; Y ) - I_\beta (X; Y )$ is nonnegative. To this end, we write this difference as: 
\begin{align}
& I (X; Y ) - I_\beta (X; Y ) \\
& = -\mathbb{E}_{X^\star Y} \log  \frac{ P^{1-\beta}(Y|X^\star) \mathbb{E}_{X}  P(Y|X)  }{ \mathbb{E}_{X}  P^\beta(Y|X) } \\
& \geq  -\log \mathbb{E}_{X^\star Y} \frac{ P^{1-\beta}(Y|X^\star) \mathbb{E}_{X}  P(Y|X)  }{ \mathbb{E}_{X}  P^\beta(Y|X) }\\
&= -\log \sum\limits_{y\in\mathcal{Y}}\mathbb{E}_{X^\star } P(y|X^\star) \frac{ P^{1-\beta}(y|X^\star) \mathbb{E}_{X}  P(y|X)  }{ \mathbb{E}_{X}  P^\beta(y|X) }\\
&=  -\log \sum\limits_{y\in\mathcal{Y}} \frac{ \mathbb{E}_{X^\star } P^\beta(y|X^\star)  \mathbb{E}_{X}  P(y|X)  }{ \mathbb{E}_{X}  P^\beta(y|X) }\\
&= -\log \sum\limits_{y\in\mathcal{Y}}  \mathbb{E}_{X}  P(y|X) \\
& = 0, 
\end{align}
where the first inequality follows by applying Jensen's inequality to the function $t\mapsto -\log (t) $. 

\textit{Proof of Proposition~\ref{th:one}:} From Lemma~\ref{lemma-inequality}, using Jensen's  inequality, we have 
\begin{align}
I (X; Y ) &=  -\mathbb{E}_{X^\star Y} \log  \mathbb{E}_{X} \left[\frac{P(Y|X)}{P(Y|X^\star)} \right], \\
&\label{eq-ineq1}\geq -\mathbb{E}_{X^\star Y} \log  \mathbb{E}_{X} \left[\frac{P(Y|X)}{P(Y|X^\star)} \right]^\beta\\
&\label{eq-ineq2}\geq -\mathbb{E}_{X^\star } \log  \mathbb{E}_{X} \mathbb{E}_{Y|X^\star } \left[\frac{P(Y|X)}{P(Y|X^\star)} \right]^\beta \\
& = -\mathbb{E}_{X^\star } \log  \mathbb{E}_{X}  \\ & \sum\limits_{y\in\mathcal{Y}} P^\beta(Y|X) P^{1-\beta}(Y|X^\star),\label{eq-ineq3}
\end{align}
where inequality (\ref{eq-ineq1}) follows by applying Lemma~\ref{lemma-inequality} and inequality (\ref{eq-ineq2}) follows by exploiting the convexity of the function $t\mapsto -\log (t) $ for any $0\leq \beta \leq 1$. Finally, it is not difficult to check from the definition of the Fisher-Rao distance given by expression~(\ref{eq:fisher_rao_multiclass}) that 
\begin{align} \label{eq:identity}
\cos & \left( \frac{d_{\textrm{FR}}( P(y|X=x) , P(y|X=x^\star))}{2} \right) =  \\ &\sum\limits_{y\in\mathcal{Y}} \sqrt{ P(y|X=x) P(y|X=x^\star)}. 
\end{align}
Using the identity given by  (\ref{eq:identity}) in expression (\ref{eq-ineq3}), and setting $\beta=1/2$, we obtain the following lower bound on $I (X; Y )$: 
$$-\mathbb{E}_{X^\star } \log  \mathbb{E}_{X} \cos \left( \frac{d_{\textrm{FR}}( P(y|X) , P(y|X^\star))}{2} \right) 
$$
The inequality (\ref{eq:kl_bound1}) immediately follows by replacing the distribution of the r.v. $X$ with the empirical distribution on the query and $P(y|x)$ with the soft-prediction corresponding to the feature $x$, which concludes the proof of the proposition.

\section{Additional Experimental Results}

\subsection{Preliminary Classification Results}

\textbf{Preliminary Experiment.} In our experiments, the backbone models are of utmost importance. Our objective in this preliminary experiment is to assess the efficacy of these models when fine-tuning \textbf{only} the model head across a variety of datasets. Through this evaluation, we aim to gain insight into their generalization abilities and any dataset-specific factors that may influence their performance. This information can be utilized to analyze the performance of different models in the few-shot scenario, as described in \autoref{sec:experiment}. We present the results of this experiment in \autoref{tab:preliminary}, noting that all classes were considered, which differs from the episodic training approach detailed in \autoref{sec:experiment}.
\begin{table*}
 \caption{Preliminary experiment results. Accuracy of the different backbone.}
    \label{tab:preliminary}
\centering
\resizebox{0.8\textwidth}{!}{\begin{tabular}{cccccccccc}\hline
       Model & Params   & Emotion  & Twitter &   Clinic & Banking  &  \multicolumn{4}{c}{Amazon}  \\\hline
               &    & en  & en &   en & en  &  en & fr & es & de \\\hline
     Albert Small V2  (XS)       &  11M  & 25.2 & 18.3 & 67.0 & 88.1   & 33.5 & X &X &X\\
     MiniLM      (S)           &  33M  & 30.2 & 19.3 & 67.1 & 92.3 &39.5  & X & X&X\\
     MPNET-base  (B)           &  109M  & 30.2 & 22.5 & 67.4 & 94.3 & 41.3  & X &X &X\\\hline
     DistilRoBERTa (S)         &  82M  & 23.3 & 26.0 & 68.5& 90.9 & 40.0  & X & X&X\\
     RoBERTa (B)         &  124M  & 21.0 & 25.5 & 66.7& 91.4 &39.2 & X &X &X\\
     RoBERTa (L)        &  355M  & 15.0 & 23.0 & 64.5& 90.0 & 38.1 & X &X&X \\\hline
     XLM-RoBERTa (B)     &  278M  & 21.0 & 22.1 & 66.5 & 87.0 & 40.1 & 19.2 &17.5 &18.3\\  
     XLM-RoBERTa (L)    &  559M  & 14.0 & 18.0 & 64.5 & 86.2 & 38.2 & 17.5 &15.6 &18.1\\  
     XLM-RoBERTa (XL)      &  3.48B   & 25.4 &19.0 & 68.9&95.0 & 41.0 & 18.9  & 17.9& 22.0\\\hline
     \texttt{text-davinci}   & 175B  & 38.9 & 35.3 & 70.4 & 98.7 & 48.4& 30.4   &34.0 & 33.5 \\\hline
    \end{tabular}}
\end{table*}

\subsection{A Dive Into \texttt{text-davinci} results}\label{ssec:dive_gpt}

\texttt{text-davinci} appears to be the backbone providing the most informative a priori embeddings in \autoref{tab:preliminary} and could be considered as the prime model for API-based FSL, showcasing the current requirements in this area. It is thus a typical candidate for application uses that must meet the following criteria {\bf (R1)} - {\bf (R3)}. Therefore, we put a special emphasis on its related results.

\begin{figure}
\centering
    \subfloat{{\includegraphics[width=0.45\textwidth] {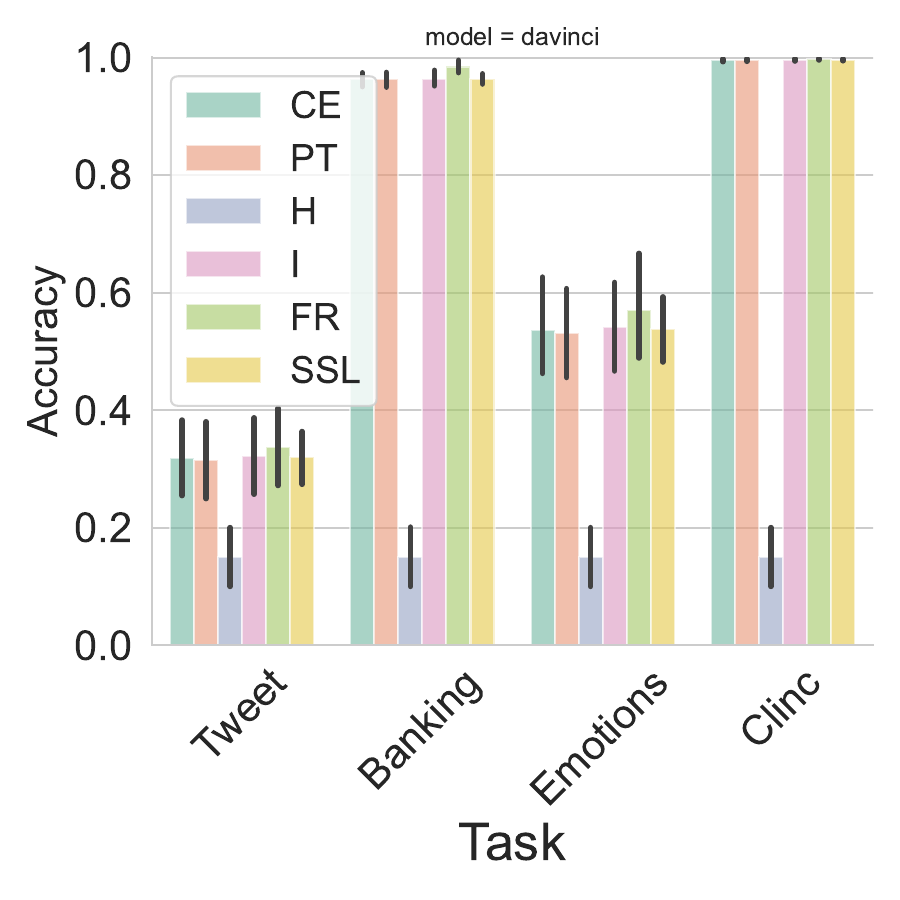}} } 
    \captionof{figure}{The different losses when training a on \texttt{text-davinci} embeddings.}
    \label{fig:gpt3_all}
\end{figure}

\autoref{fig:gpt3_all} (top) details the \texttt{text-davinci} results of the experiments conducted on the mono-lingual datasets. These plots highlight the consistency of the tendencies that emerged in \autoref{tab:preliminary}, \autoref{tab:global} and \autoref{fig:monolingual}, namely: the superiority of transductive approaches ($FR$ and $I$) over inductive ones ($CE$ and $PT$), the underperformance of the entropic-minimization-based strategy (H), and the higher amount of information conveyed by \texttt{text-davinci} learned embeddings over other backbones, resulting in higher F1 scores on all datasets.

These phenomena still occur in the multi-lingual setting, as illustrated in \autoref{fig:gpt3_all} (bottom), stressing the superiority of transductive (and especially FR) over other approaches for presumably universal tasks, beyond English-centered ones, and without the need for using language-specific engineering as for prompting-based strategies.

Note that for both of these settings, the entropic-minimization-based strategy (H) seems to be capped at a 15\% F1 score, thus with no improvement over other backbones embeddings, and independently of the dataset difficulty. 

\subsection{Multilingual Experiment}\label{ssec:additionnal_multilingua}

To provide an exhaustive analysis, we report the same experiment that is made in \autoref{subsec:global}, for multi-lingual models on the Amazon dataset.
While both Latin languages (French and Spanish) share close results, with an F1 gain of 2.8\% for FR over CE, the results in the German and English language exhibit an F1 increased by almost 4\%.   
\begin{figure*}
  \centering \subfloat{{\includegraphics[width=7.5cm]{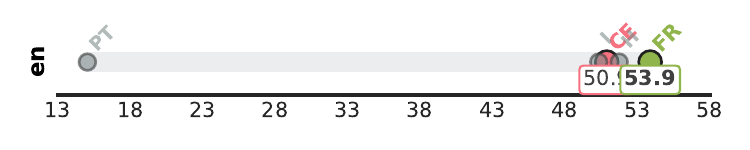} }} \\ \subfloat{{\includegraphics[width=7.5cm]{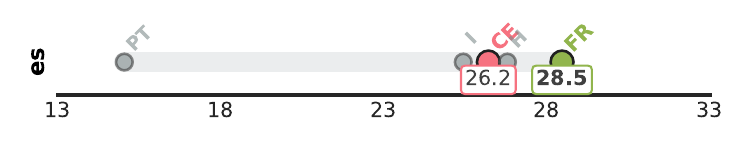} }} \\ \subfloat{{\includegraphics[width=7.5cm]{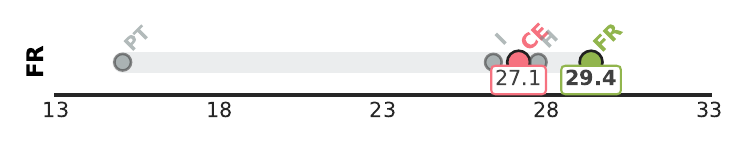} }} \\ 
  \subfloat{{\includegraphics[width=7.5cm]{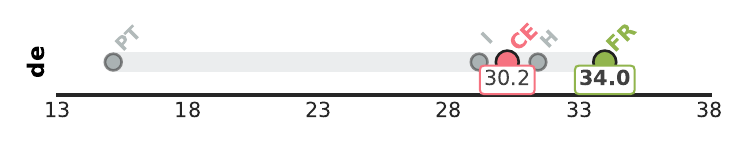} }}
    \captionof{figure}{Losses Performance on multilingual datasets.}%
    \label{fig:all_dynamical}%
\end{figure*}

\subsection{Importance of Model Backbones on Monolingual Experiment}
In this section, we report the results of our experiment aggregated per backbone. The goal is to understand how the different losses behave on the different backbones. The results are presented in \autoref{fig:all_dynamical}. While the trends observed in the previous charts are retrieved for the majority of backbones, some of these models are exceptions. For example, while transductive methods perform generally better than inductive methods, the CE-based method seems to perform slightly better than I for XLM-RoBERTa-xl. Additionally, while FR is the most effective method for the majority of backbones, it is surpassed by I for the all-distilroberta-v1 model.
Furthermore, the inverse-scaling-law details are found for the RoBERTa(B/L) and XLM-RoBERTa
(B/L) models per dataset. In general, it is interesting to note that although model performance is constrained by dataset difficulty, the performance order of each method is consistent across all 4 datasets for each considered backbone.

\begin{figure*}[!ht]
    \centering
    \subfloat{{\includegraphics[width=12cm]{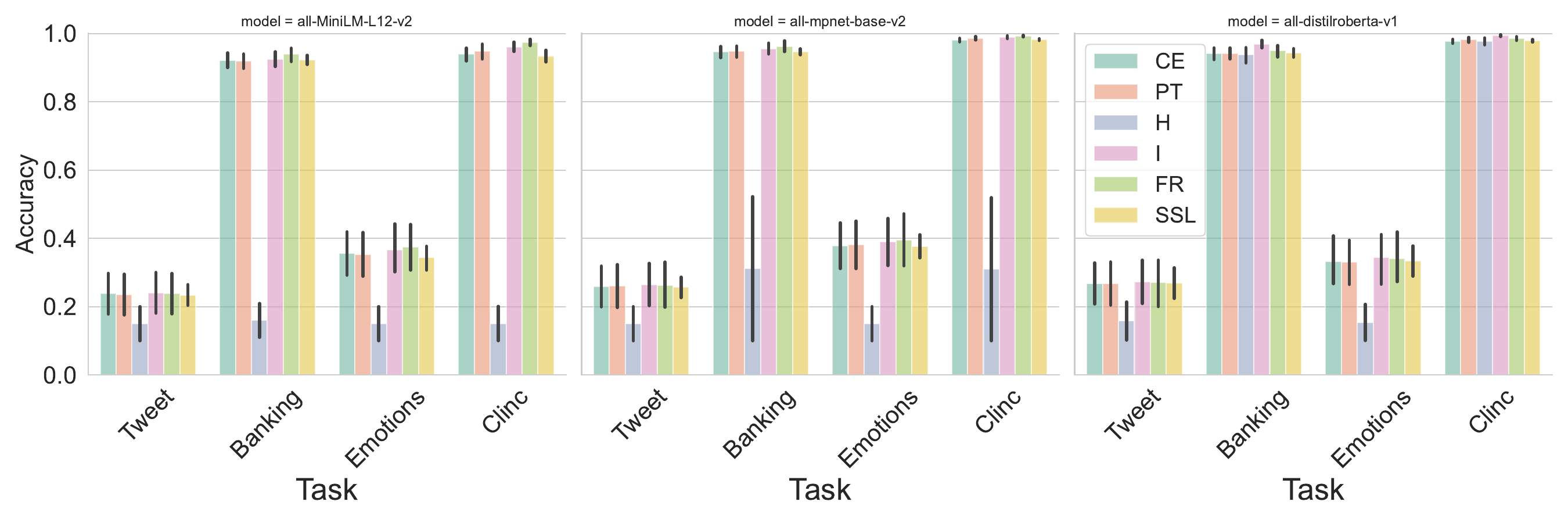} }} \\
        \subfloat{{\includegraphics[width=12cm]{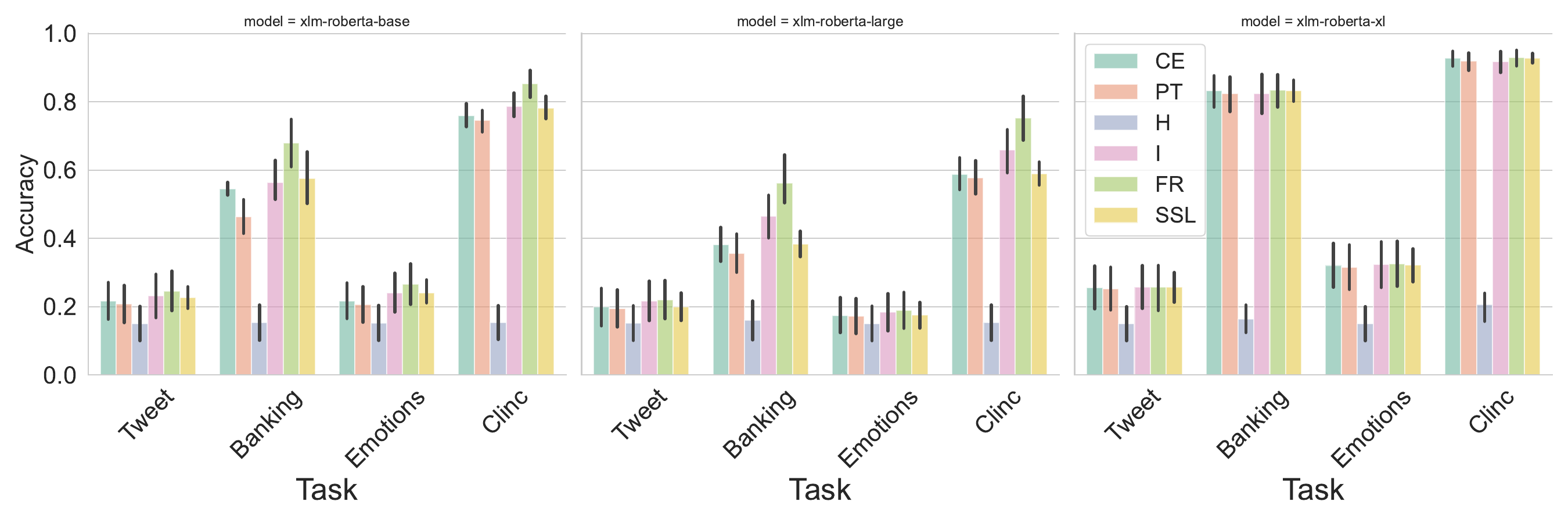} }} \\
            \subfloat{{\includegraphics[width=12cm]{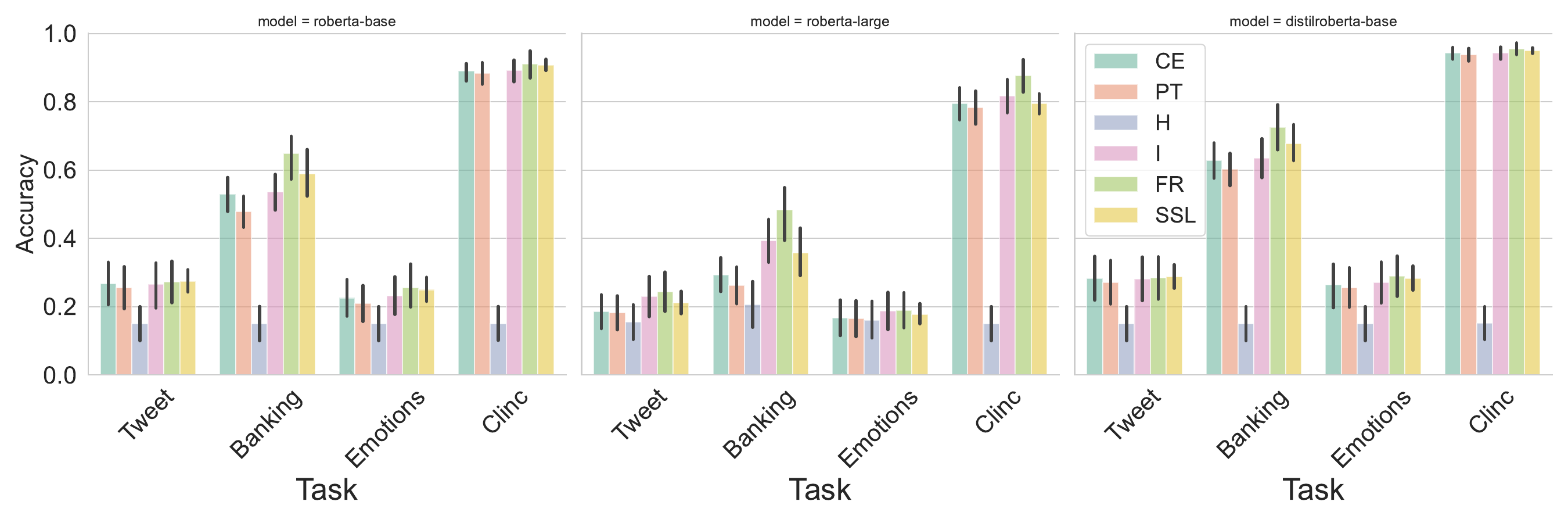} }}  \\
    \captionof{figure}{Performance on monolingual datasets.}%
    \label{fig:all_dynamical}%
\end{figure*}

\subsubsection{Results Per Language}
In this experiment, we report the performance of different losses on the Amazon dataset by averaging the results over the number of shots, ways, and model backbones. The results are presented in \autoref{tab:mulitlingualk_all_over}. Our observations indicate that the transductive regularization improves the results for all languages over the inductive baseline (i.e., CE), with a substantially higher gain for the German language. Additionally, we note that the observed improvements for FR are more consistent. This further demonstrates that the transductive loss can be useful in few-shot NLP. In the future, we would like to explore the application of transductive inference to other NLP tasks such as sequence generation \cite{pichler2022differential,colombo-etal-2019-affect,colombo-etal-2021-novel,colombo-etal-2021-improving} and classification tasks \cite{chapuis-etal-2020-hierarchical,colombo-etal-2022-learning,colombo2022best,himmi2023towards} as well as NLG evaluation \cite{colombo-etal-2021-automatic,colombo2022glass,colombo-etal-2021-beam,colombo-etal-2021-code,colombo-etal-2021-improving} and Safe AI \cite{colombo2022beyond,picot2022adversarial,picot2022simple,darrin2022rainproof,darrin2023unsupervised}.
\begin{table}[]
    \centering
\begin{tabular}{lrrrr}\hline
  & fr  &  de   & en   & es \\\hline
  FR     &  \textbf{29.36} &  \textbf{33.98} &  \textbf{53.89} &  \textbf{28.47} \\
  I      &  \underline{27.74} &  \underline{31.41} &  \underline{51.75} &  \underline{26.79} \\
  H      &  15.04 &  15.13 &  15.04 &  15.04 \\\hline
CE     &  27.15 &  30.24 &  50.89 &  26.21 \\
PT     &  26.37 &  29.16 &  50.34 &  25.44 \\
SSL  & 27.20	&30.29	&50.94	&26.26 \\\hline
\end{tabular}
    \caption{Global Results for multilingual Amazon}
    \label{tab:mulitlingualk_all_over}
\end{table}

\end{document}